\documentclass{article} % For LaTeX2e
\usepackage{iclr2025_conference,times}

\usepackage{amsmath,amsfonts,bm}

\def\eqref#1{equation~\ref{#1}}
\def\1{\bm{1}}

\def\ra{{\textnormal{a}}}

\DeclareMathAlphabet{\mathsfit}{\encodingdefault}{\sfdefault}{m}{sl}
\SetMathAlphabet{\mathsfit}{bold}{\encodingdefault}{\sfdefault}{bx}{n}

\newcommand{\E}{\mathbb{E}}

\newcommand{\R}{\mathbb{R}}

\newcommand{\Var}{\mathrm{Var}}

\newcommand{\Cov}{\mathrm{Cov}}
\usepackage{graphicx}
\usepackage{booktabs} % for toprule
\usepackage{hyperref,comment}
\usepackage{url}
\usepackage{enumerate,enumitem,wrapfig}
\usepackage{multirow,multicol}
\usepackage{subfigure}
\usepackage{amsmath}
\usepackage{mathtools}
\usepackage{microtype}
\title{TGB-Seq Benchmark: Challenging Temporal GNNs with Complex Sequential Dynamics}

\author{Lu Yi\textsuperscript{1}, Jie Peng\textsuperscript{1}, Yanping Zheng\textsuperscript{1}\thanks{Yanping Zheng and Zhewei Wei are the corresponding authors.}, Fengran Mo\textsuperscript{2}, Zhewei Wei\textsuperscript{1}\footnotemark[1] \\
\textsuperscript{1}Renmin University of China, \textsuperscript{2}Université de Montréal\\
\texttt{\{yilu, peng\_jie, zhengyanping, zhewei\}@ruc.edu.cn} \\
\texttt{fengran.mo@umontreal.ca}
\And
Yuhang Ye\textsuperscript{3}, Zixuan Yue\textsuperscript{3} \\
\textsuperscript{3}Huawei Poisson Lab, Huawei Technology Ltd. \\
\texttt{\{yeyuhang,yuezixuan\}@huawei.com}
\And  
Zengfeng Huang\textsuperscript{4} \\
\textsuperscript{4}Fudan University, Shanghai Innovation Institute \\
\texttt{huangzf@fudan.edu.cn} 
}% The \author macro works with any number of authors. There are two commands
\iclrfinalcopy % Uncomment for camera-ready version, but NOT for submission.
\begin{document}

\maketitle

\newtheorem{fact}{Fact}
\newtheorem{observation}{Observation}
\def\header{\vspace{0.8mm} \noindent}

\def\review{\vspace{3mm} \noindent}

\def\algocapup{\vspace{-4mm}}
\def\algocapdown{\vspace{-4mm}}
\def\tblcapup{\vspace{0mm}}
\def\tblcapdown{\vspace{1mm}}
\def\tbldown{\vspace{-0mm}}
\def\figcapup{\vspace{-2mm}}
\def\figcapdown{\vspace{-2mm}}
\def\theoremtop{\vspace{1mm}}
\def\theoremdown{\vspace{1mm}}

\newcommand{\pushright}[1]{\ifmeasuring@#1\else\omit\hfill$\displaystyle#1$\fi\ignorespaces}
\newcommand{\pushleft}[1]{\ifmeasuring@#1\else\omit$\displaystyle#1$\hfill\fi\ignorespaces}
\newcommand{\vit}[1]{{\color{red} #1}}
\newcommand{\try}[1]{{\color{blue} [ #1 ]}}
\newcommand{\redrev}[1]{{\color{red} #1}}

\def\la{\langle}
\def\ra{\rangle}

\newcommand{\eqn}[1]{{Equation~(\ref{#1})}}
\newcommand{\reqn}[1]{{Equation~(#1)}}
\newcommand{\ineqn}[1]{{Inequality~(\ref{#1})}}
\newcommand{\rineqn}[1]{{Inequality~(#1)}}
\newcommand{\alg}[1]{{Algorithm~\ref{#1}}}
\newcommand{\ralg}[1]{{Algorithm~{#1}}}
\newcommand{\thm}[1]{{Theorem~\ref{#1}}}
\newcommand{\term}[1]{{Term~(\ref{#1})}}
\newcommand{\nodef}[1]{{{\bf{x}}_{#1}}}
\newcommand{\edgef}[2]{{{\bf{e}}_{#1,#2}}}
\newcommand{\mem}[0]{{\bf mem}}

\def\appG{\hat{G}}
\def\aggfunc{{\mbox{AGGR}}}
\def\memfunc{\mbox{MEM}}
\def\emb{{\bf emb}}
\def\CoNeifunc{{\mbox{CO-REL}}}
\def\rel{{\bf rel}}
\def\Ex{\mathrm{E}}
\def\Var{\mathrm{Var}}
\def\Cov{\mathrm{Cov}}
\newcommand{\lutodo}[1]{{\color{blue} [Lu todo: #1]}}
\newcommand{\jietodo}[1]{{\color{yellow} [Jie todo: #1]}}
\newcommand{\pingtodo}[1]{{\color{orange} [Ping todo: #1]}}
\newcommand{\notsure}[1]{{\color{red} [#1]}}
\def\n{n}
\def\d{\bar{d}}

%\def\I{\mathcal{I}}
%\def\Pcal{\mathcal{I}}
%\def\s{\mathbf{s}}
%\def\t{\mathbf{t}}
% methods
\def\G{\mathcal{G}}
\def\e{\mathbf{e}}
\def\V{\mathcal{V}}
\def\E{\mathcal{E}}
\def\loss{l}
\def\adj{\mathbf{A}}
\def\tadj{\tilde{\mathbf{A}}}
\def\deg{\mathbf{D}}
\def\tdeg{\tilde{\mathbf{D}}}
\def\feat{\mathbf{X}}
\def\featvec{\mathbf{x}}
\def\R{\mathbb{R}}
\def\Rsd{\mathbf{R}}
\def\rsum{\r_{\rm sum}}
\def\W{\mathbf{W}}
\def\w{\mathbf{w}}
\def\prop{\mathbf{P}}
\def\CloseFloat{\setlength{\textfloatsep}{2.5mm}}
\def\OpenFloat{\setlength{\textfloatsep}{16pt}}
\def\step{L}
\def\newemb{\mathbf{Z}'}
\def\Q{\boldsymbol{q}}
\def\q{\boldsymbol{q}}
\def\appemb{\mathbf{\hat{Z}}}
\def\appnewemb{\mathbf{\hat{Z}'}}
\def\embvec{\mathbf{z}}
\def\appembvec{\mathbf{\hat{z}}}
\def\newprop{\mathbf{P}'}
\def\Loss{\mathcal{L}}
\def\Lossb{\mathcal{L}_\mathbf{b}}
\def\appLoss{\hat{\mathcal{L}}}
\def\appLossb{\hat{\mathcal{L}}_\mathbf{b}}
\def\appD{\hat{\mathcal{D}}}
\def\I{\mathcal{I}}
\def\b{\mathbf{b}}
\def\ldeg{\mathbf{d}}
\def\rmax{r_{\max}}
\def\T{\boldsymbol{T}}
\def\t{\boldsymbol{t}}
\def\AL{\mathcal{A}}
\def\appAL{\mathcal{A}}
\def\AUL{\mathcal{M}}
\def\D{\mathcal{D}}
\def\Diag{\mathbf{D}}
\def\P{\mathbf{P}}
\def\H{\mathcal{H}}
\def\Hes{\mathbf{H}}
\def\Y{\mathbf{Y}}
\def\y{\mathbf{y}}
\def\appHes{\hat{\Hes}}
\def\nei{\mathcal{N}}
\def\one{\mathbf{1}}
\def\u{\mathbf{u}}
\def\w{\mathbf{w}}
\def\optw{\mathbf{w}^\star}
\def\U{\mathbf{U}}
\def\W{\mathbf{W}}
\def\A{\mathbf{A}}
\def\B{\mathbf{B}}
\def\C{\mathbf{C}}
\def\zero{\mathbf{0}}
\def\T{\mathbf{T}}
\def\appT{\hat{\mathbf{T}}}
\def\apppi{{\hat{\boldsymbol{\pi}}}}
\def\ppi{{\boldsymbol{\pi}}}
\def\r{\boldsymbol{r}}
\def\Res{\boldsymbol{R}}

\newenvironment{proofm}
{\par\medskip\indent{\bf\upshape Proof}\hspace{0.5em}\ignorespaces}
{\hfill\par\medskip}

%%% Local Variables:
%%% mode: latex
%%% TeX-master: "paper"
%%% End:

\begin{abstract}
Future link prediction is a fundamental challenge in various real-world dynamic systems. To address this, numerous temporal graph neural networks (temporal GNNs) and benchmark datasets have been developed. 
However, these datasets often feature excessive repeated edges and lack complex sequential dynamics, a key characteristic inherent in many real-world applications such as recommender systems and ``Who-To-Follow'' on social networks. This oversight has led existing methods to inadvertently downplay the importance of learning sequential dynamics, focusing primarily on predicting repeated edges.

In this study, we demonstrate that existing methods, such as GraphMixer and DyGFormer, are inherently incapable of learning simple sequential dynamics, such as ``a user who has followed OpenAI and Anthropic is more likely to follow AI at Meta next.'' Motivated by this issue, we introduce the \underline{T}emporal \underline{G}raph \underline{B}enchmark with \underline{Seq}uential Dynamics (TGB-Seq), a new benchmark carefully curated to minimize repeated edges, challenging models to learn sequential dynamics and generalize to unseen edges. TGB-Seq comprises large real-world datasets spanning diverse domains, including e-commerce interactions, movie ratings, business reviews, social networks, citation networks and web link networks. Benchmarking experiments reveal that current methods usually suffer significant performance degradation and incur substantial training costs on TGB-Seq, posing new challenges and opportunities for future research. TGB-Seq datasets, leaderboards, and example codes are available at \url{https://tgb-seq.github.io/}.
\end{abstract}

\section{Introduction}\label{sec:intro}
Future link prediction~\citep{future_link_prediction} is a fundamental challenge in various real-world dynamic systems, such as social networks~\citep{app_social}, e-commerce~\citep{bai2020temporal}, financial systems~\citep{financial}. For instance, an online shopping website must decide which items to recommend to users based on their click history, while a social networking platform needs to identify which users may be interested in connecting based on their existing relationships.
Among the various approaches for future link prediction, temporal Graph Neural Networks (GNNs) are particularly notable for their flexibility in modeling diverse applications and their representation learning capabilities~\citep{zheng2024survey,Skarding_survey,mach_survey}. Recently, several temporal GNN methods~\citep{DyGFormer} have demonstrated impressive performance in future link prediction on existing benchmarks~\citep{edgebank}. However, most existing datasets are not derived from real-world recommender systems, despite recommendations being a natural and essential application of future link prediction. 

\header{\bf Observations.} To assess the capability of current temporal GNNs in recommendation tasks, we evaluate their performance on future link prediction using {two widely used recommendation datasets, including 
% the movie rating network ML-20M~\footnote{https://grouplens.org/datasets/movielens/20m}, 
the user-product interaction network Taobao~\citep{Taobao1} and the business review network Yelp~\footnote{https://www.yelp.com/dataset}. 
% and GoogleLocal~\citep{googlelocal1,googlelocal2}.}
Figure~\ref{fig:rec} presents the performance of three state-of-the-art temporal GNN approaches across these datasets, including EdgeBank~\citep{edgebank}, GraphMixer~\citep{GraphMixer} and DyGFormer~\citep{DyGFormer}.
% Following the evaluation protocol in~\citet{TGB}, 
We split these datasets chronologically and randomly sample 100 negative destination nodes for each positive instance, utilizing the Mean Reciprocal Rank (MRR) as the evaluation metric. Besides, we also include SGNN-HN~\citep{SGNNHN},
\begin{wrapfigure}{r}{0.5\textwidth}
  \centering
  \hspace{-4mm}
  \resizebox{\linewidth}{!}{
  \begin{tabular}{c}
  \includegraphics[width=55mm]{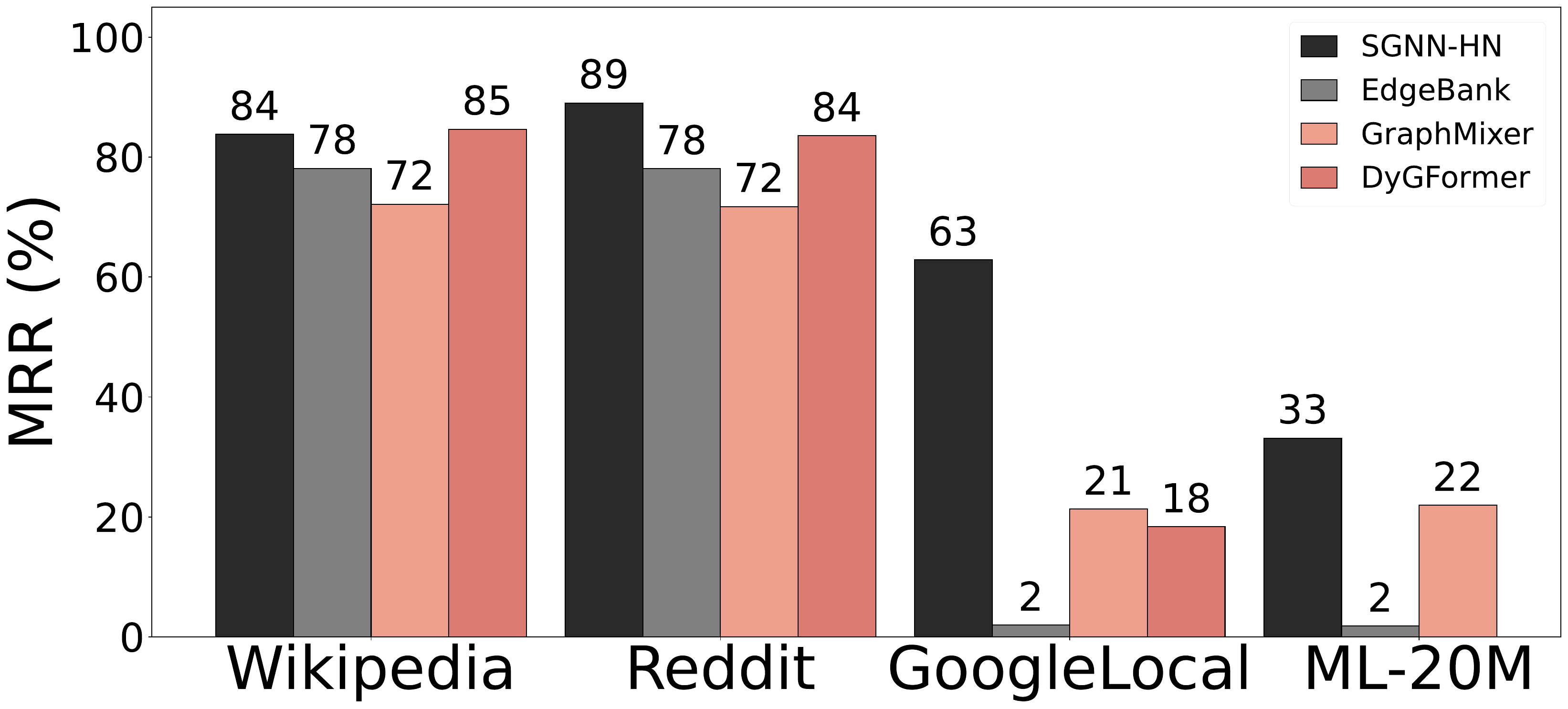}
  % \hspace{-2mm} 
  % \includegraphics[width=40mm]{./fig/intro_scale.pdf} 
  % \hspace{-2mm} \includegraphics[width=44mm]{./06-03/exp_1_100000_1.pdf} &
  % \includegraphics[width=44mm]{./06-03/lognormal_100000_1.pdf}
  \hspace{-4mm} 
  \end{tabular}
  }
  \vspace{-4mm}
  \caption{The MRR scores of three selected temporal GNNs and SGNN-HN on two existing datasets (Wikipedia, Reddit) and two recommendation datasets (Yelp and Taobao). }
  \label{fig:rec}
  \vspace{-2mm}
\end{wrapfigure}
one of the state-of-the-art methods for sequential recommendation to compare with temporal GNNs.
% Detailed information about the datasets and experimental settings are provided in Section~\ref{sec:datasets} and~\ref{sec:exp}.
% Table~\ref{tbl:rec} presents the performance of nine popular methods on the four recommendation datasets. 
%include the performance of SGNN-HN~\citep{SGNNHN}, one of the state-of-the-art sequential recommendation methods, for comparison. 
Intuitively, these recommendation datasets are comparable to existing datasets (e.g., Wikipedia and Reddit), as all represent typical dynamic systems, and thus, temporal GNNs are expected to perform in a similar trend on these recommendation datasets.
However, Figure~\ref{fig:rec} shows that temporal GNN methods present significant performance degradation compared to their strong results on two previously established datasets and present a substantial performance gap compared with SGNN-HN, which contradicts our intuition.
%the results reveal that temporal GNNs demonstrate significant performance degradation compared to their strong results on established datasets and maintain a substantial performance gap relative to SGNN-HN.
%Surprisingly, the temporal GNN methods underperform on these recommendation datasets, exhibiting a significant performance gap compared to the specialized recommendation method SGNN-HN. 
This arises a critical question: 
\textit{Why do existing temporal GNN methods, which demonstrate superior performance on existing temporal graph datasets, fail to perform well in a typical downstream application, i.e., recommendation?}

%\textit{Why do existing temporal GNNs demonstrate superior performance on established temporal graph datasets but underperform in typical recommendation scenarios?}

%We observe that existing datasets contain significantly more repeated edges compared to these recommendation datasets. Consequently, temporal GNNs can readily predict these repeated edges, resulting in high performance on established benchmarks.
We conjecture that this is because existing datasets, e.g., Wikipedia and Reddit, contain excessive repetitions of historical edges compared to these evaluated recommendation datasets.
Consequently, temporal GNNs tend to predict these repeated historical edges via memorizing or aggregating historical edges and perform well on existing datasets.
To validate our assumption, 
%we assess the capabilities of current methods in predicting both 
we use existing temporal GNNs to predict both repeated and unseen edges and report their MRR scores separately across four widely used datasets: Wikipedia, Reddit, Social Evo. and Enron, following the experimental settings of Figure~\ref{fig:rec}.
%replicate those of Table~\ref{tbl:rec}. 
% Figure~\ref{fig:hist} shows the MRR score for test edges that are repetitions of historical edges and unseen edges, where the latter are denoted by ``Unseen''. 
% Edgebank~\citep{edgebank} always achieves an MRR of 1.0 for historical edge predictions and 0.0 for unseen edges, and thus its results are omitted. 
%for test samples that are repetitions of historical edges and the MRR for test samples that are unseen edges, denoted by ``NonHist''. 
% The results 
The results in Figure~\ref{fig:hist}
indicate a substantial prediction performance gap between historical and unseen edges, with differences reaching up to eightfold. This phenomenon implies that existing methods are effective on graphs dominated by repeated edges but fail to generalize to those that emphasize unseen edges. 
%excel at predicting repeated edges but struggle to generalize to unseen edges. 
The underlying reason is probably that existing methods tend to rely on the information of historical neighbors, which limits their generalizability.
Thus, they can only associate query nodes with their historical neighbors but fail on unseen edges.

%for this limitation appears to be 
%the methods' reliance on memorizing and aggregating information from historical neighbors. 
%While this approach effectively associates query nodes with their historical edges, it falls short in generalizing to unseen edges. 
%This limitation hampers the models' ability to perform robustly in scenarios where predicting novel interactions is crucial, such as recommendation.

% Aiming to predict unseen interactions, the models must capture complex 
% which are crucial features in real-world dynamic systems. 
\header{\bf Motivations.} However, future links are typically not simple repetitions of historical ones in many real-world dynamic systems. Instead, the evolution of many dynamic systems often exhibits intricate \textit{sequential dynamics}. For example, on an e-commerce platform, an \textit{entity} (i.e., a customer) who has purchased a smartphone and a phone case is likely to buy a screen protector next. In this context, future interactions of entities typically involve new purchases rather than simply repeating past ones.
Therefore, a model must capture the inherent sequential dynamics in these systems to accurately predict future links. \textit{Capturing sequential dynamics involves modeling the evolution of the intentions of entities based on their historical interactions and forecasting unseen interactions. }
However, we find that existing temporal GNNs struggle to effectively capture even simple sequential dynamics that exclude repeated edges, despite these dynamic patterns being frequently present in the training set. The observed cases are provided in later Section~\ref{sec:pitfalls} and Figure~\ref{fig:example}.
On the other hand, existing datasets often contain an excessive number of repeated edges, which undermines the critical aspect of complex sequential dynamics. Evaluating temporal GNN models solely based on these datasets cannot adequately assess their ability to capture complex sequential dynamics.
%as demonstrated in Section~\ref{sec:pitfalls} and Figure~\ref{fig:example}, current temporal GNNs fail to effectively capture simple interaction patterns, even when these patterns are repeatedly present in the training set. 
% such as a user who interacts sequentially with items $\{i_0, i_1, i_2,i_3\}$ will definitely interact with items $i_4$ next.
%existing datasets often contain an excessive number of repeated edges, thereby neglecting the importance of complex sequential dynamics in real-world applications. This limitation renders them inadequate for comprehensively evaluating the capabilities of models in capturing sequential dynamics.
%  This limitation hinders temporal GNNs from making accurate predictions in real-world applications such as recommendations.
% An effective prediction models must learn the evolution of entities' intentions based on their historical interactions and generalize to unseen interactions. 
\begin{figure}[t]
  \begin{minipage}[t]{1\textwidth}
  \centering
  % \vspace{1mm}
  \begin{tabular}{c}
  % \hspace{-22mm} 
  \includegraphics[width=136mm]{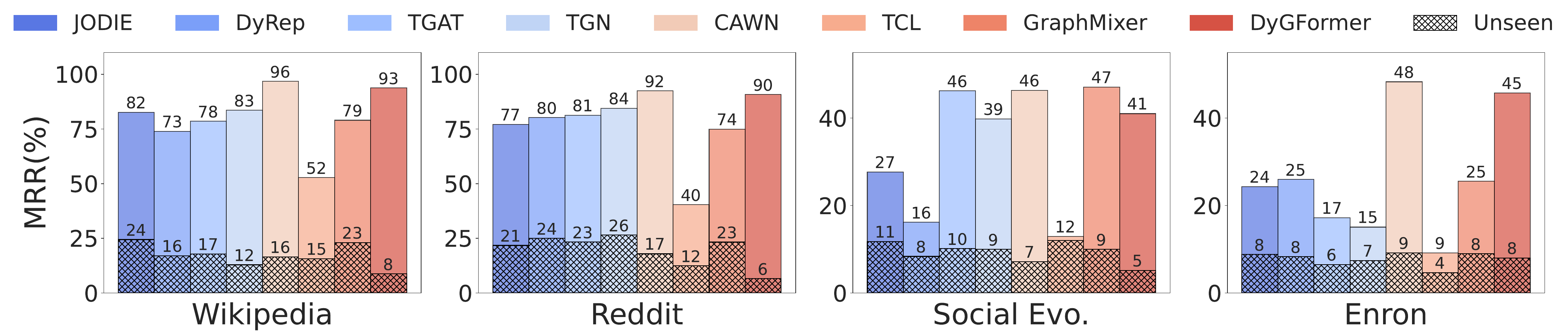}
  \end{tabular}
  \vspace{-4mm}
  \caption{The MRR scores of eight popular temporal GNNs for predicting repeated historical edges on four previously established datasets. ``Unseen'' denotes the performance of unseen edges.}
  \label{fig:hist}
  \vspace{-3mm}
  \end{minipage}
\end{figure}

\header{\bf Contributions.} To address this gap, we present the \underline{T}emporal \underline{G}raph \underline{B}enchmark with \underline{Seq}uential Dynamics (TGB-Seq), a collection of new benchmark datasets designed to evaluate %existing methods' 
the systems' ability to capture complex sequential dynamics. TGB-Seq includes four widely-used recommendation datasets and four non-bipartite datasets derived from typical future link prediction scenarios that inherently exhibit complex sequential dynamics, including a movie rating network (ML-20M), an e-commerce interaction network (Taobao), two business review networks (Yelp and GoogleLocal), two ``Who-To-Follow'' social networks (Flickr and YouTube), a citation network (Patent) and a web link network (WebLink). 
%These datasets comprise two ``Who-To-Follow'' datasets, a citation network and a web link network.
TGB-Seq datasets are carefully curated to minimize repeated edges. Only Yelp and Taobao contain a small number of repeated edges with the natural behavior that users potentially review or click items multiple times. 
All TGB-Seq datasets are ensured with medium to large scale toward the practical situation, %of medium to large scale,
comprising millions to tens of millions of edges. 
%Upon evaluating existing temporal GNNs on TGB-Seq, we find that these methods consistently experience significant performance degradation compared to their impressive results on established datasets.
% Furthermore, most temporal GNNs prove inefficient on large-scale datasets, incurring high training costs. 
%when applied to TGB-Seq datasets and underscore the importance of TGB-Seq in evaluating models' abilities to capture complex sequential dynamics.
% provide a new framework for evaluating temporal GNNs and present fresh challenges for future research, encouraging the development of more robust and efficient temporal graph neural networks.
Overall, we make the following contributions in this paper:
\begin{itemize}[leftmargin = *]
    \item We demonstrate that existing temporal GNNs fail to capture sequential dynamics in temporal graphs, limiting their generalizations to various real-world scenarios. 
    \item We propose TGB-Seq, a collection of eight benchmark datasets for future link prediction, carefully curated from diverse application domains with intricate sequential dynamics. 
    TGB-Seq focuses on evaluating temporal GNNs' capability to capture sequential dynamics and generalize to unseen edges, addressing the limitations of existing datasets that contain excessive repetitions of edges and overlook the intricate sequential dynamics present in real-world dynamic systems.
    \item Comprehensive evaluations on TGB-Seq reveal that existing temporal GNNs experience substantial performance declines compared to their impressive results on existing benchmarks. This observation underscores the limited ability of existing methods to capture complex sequential dynamics and demonstrate the distinguishing functionality of TGB-Seq in evaluating such ability.
    \item We provide a Python package available via pip, enabling seamless dataset downloading, negative sample generation, and evaluation. All code is publicly available on the \href{https://github.com/TGB-Seq/TGB-Seq}{TGB-Seq GitHub repository}. Researchers can submit their methods and compare performance on the \href{https://tgb-seq.github.io/}{TGB-Seq website}, where we maintain the leaderboards for all datasets.
\end{itemize}

\section{Related Work}
% Jodie, EdgeBank, TGB, TGB2.0, DTGB(Dynamic Text-attributed Graph Benchmark), A Robust Comparative Analysis of Graph Neural Networks on Dynamic Link Prediction,
\header{\bf Temporal Graph Datasets and Benchmarks.}
Several studies~\citep{edgebank,TGB} have highlighted that existing benchmarks for dynamic graph learning often lead to overly optimistic assessments of current approaches. Specifically, widely used datasets such as Reddit, Wikipedia, MOOC, and LastFM~\citep{Jodie} suffer from inconsistent preprocessing and simplistic negative sampling strategies, which result in inflated performance metrics and unreliable comparisons. To address these issues, BenchTeMP~\citep{Benchtemp} provides a unified evaluation framework with consistent datasets and comprehensive performance metrics. \citet{edgebank} construct six dynamic graph datasets across diverse domains, such as politics, economics, and transportation, and introduce two negative sampling strategies to create more challenging evaluations. 
TGB~\citep{TGB} further advances the field by introducing several large datasets for future link prediction, establishing a comprehensive benchmark based on the unified evaluation metric of MRR.
~\citet{gastinger2024tgb2} extend TGB with multi-relational datasets for Temporal Knowledge Graphs (TKG) and Temporal Heterogeneous Graphs (THG). These extensions focuses on future link prediction on large-scale TKG and THG datasets.
% Different from previous work that emphasizes expanding dataset diversity, scale, and evaluation protocols, we construct a collection of challenging benchmark datasets that originate from typical application scenarios of future link prediction, challenging the existing temporal GNNs with complex sequential dynamics inherently in these applications.
%we focus on analyzing the characteristics of existing datasets and constructing more complex ones that originate from typical application scenarios of future link prediction.

\header{\bf Temporal Graph GNNs for future link prediction.} % dtdg, ctdg
Future link prediction is a critical task in various dynamic systems, which aim to predict future interactions or relationships between entities based on historical data. 
% Early research treated temporal graphs as a series of snapshots, applying static GNNs to encode them and using sequence analysis to capture temporal changes~\citep{GRNN, xu2019generative, Evolvegcn, Dysat, ROLAND}. 
% However, recent studies take continuous-time models as a paradigm, which view the input as a stream of interactions to preserve the evolution of temporal graphs. 
% In this paper, we also focus on the continuous-time dynamic graphs, as it better reflects how dynamic graphs form incrementally in real-world scenarios. 
To capture the evolution pattern, memory-based methods such as TGN~\citep{TGN}, Jodie~\citep{Jodie}, DyRep~\citep{DyRep}, and APAN~\citep{APAN}, use dynamic memory modules to store and update node information during interactions, allowing for more effective modeling. 
On the other hand, approaches like TGAT~\citep{TGAT}, CAWN~\citep{CAWN}, TCL~\citep{TCL}, GraphMixer~\citep{GraphMixer}, and DyGFormer~\citep{DyGFormer}, aggregate historical neighbor information directly during prediction without memory modules. These methods employ contrastive learning and Transformer-based techniques to capture evolving node interactions and temporal dependencies. The Hawkes process~\citep{Hawkes1,Hawkes2} is another widely used technique for capturing the impacts of historical events on current events and is employed by methods such as TREND~\citep{TREND} and LDG~\citep{LDG}, etc.
% utilizes the Hawkes process to model the exciting effects between sequential interactions and captures both individual and collective characteristics of events by integrating event and node dynamics.
Drawing inspiration from natural language processing (NLP) studies, SimpleDyG~\citep{simpledyg} models dynamic graphs as a sequence modeling problem, using a simple Transformer architecture without complex modifications.
\citet{edgebank} observe that edges reoccur over time in the existing datasets and propose a simple memory-based heuristic approach, EdgeBank, without any learnable components. 
This method predicts edges based solely on past observations, yet it demonstrated remarkable performance in current evaluations. 
This further highlights the need for more comprehensive benchmarks that assess models' ability to generalize to unseen edges, thereby ensuring robust performance in real-world scenarios.

\section{Task Formulation and Current Pitfalls}\label{sec:pitfalls}
Temporal graphs represent entities in the dynamic systems as nodes and interactions among entities as edges. 
% The entities and the interaction among them are represented as nodes and edges in temporal graphs.
Each edge is labeled with a timestamp to indicate the time of interaction occurred. 
Existing studies mainly categorize temporal graphs into two types: continuous-time temporal graphs and discrete-time temporal graphs. In this paper, we focus on continuous-time temporal graphs since they better reflect how dynamic graphs form incrementally in real-world scenarios and discrete-time temporal graphs can be directly converted to continuous-time temporal graphs without information loss.
Formally, a continuous-time temporal graph can be denoted as $\G=(\V,\E)$, where the edge set $\E$ can be represented as a stream of timestamped edges, i.e., $\E=\{(s_0,d_0,t_0),(s_1,d_1,t_1),\cdots,(s_T,d_T,t_T)\}$ with $s_i,d_i\in\V$ representing the source and destination nodes, respectively.
The $t_i$ denotes the timestamp of the $i$-th edge with $t_0\le t_1\le \cdots\le t_T$. 
\subsection{Future Link Prediction Formulation and Evaluation}\label{sec:evaluations}

\header{\bf Future Link Prediction.} %In current literature, the task of future link prediction is often formulated as predicting the existence of a link between two nodes at a given timestamp. 
The task of future link prediction is formulated as predicting the existence of a link between two nodes at a given timestamp in literature~\citep{Jodie}.
Specifically, given a temporal graph $\G$, a query edge $(s,d,t)$, and all edges appeared before time $t$, the model is required to predict the likelihood of the edge $(s,d)$ appearing at time $t$. 
However, in real-world applications, the fundamental objective is to determine which entities the query entity is most likely to interact with. For instance, in the “Who-To-Follow” scenario within social networks, the task is to predict which users the query user is likely to follow next. The users with the highest predicted likelihood are then recommended to the query user.
Given the high computational costs associated with calculating the likelihood of all potential entities in a large-scale graph, current literature in recommendation and knowledge graphs~\cite{NCF,SASRec,knowledgegraph} treats the future link prediction task as a ranking problem among multiple negative samples. Specifically, given a query edge $(s,d,t)$, the model needs to rank the positive destination node $d$ higher among the sampled $k$ negative destinations based on the likelihood.  
The current temporal graph benchmark study, TGB~\citep{TGB}, adopts these settings and sets $k$ to 20.
We set $k$ to 100 for a more robust evaluation in our experimental setup.
%ranking the likelihood of all potential entities is computationally expensive for large-scale datasets and thus infeasible for research.
%To balance between the current task formulation, real-world applications, and feasibility, 
%we sample 100 negative destination nodes and ask the model to rank the likelihood of the positive destination node $d$ among the negative destinations. %This setting is also widely-adopted in recommendation literature~\citep{NCF,SASRec} and knowledge graphs~\citep{knowledgegraph}. 
% Recently, multiple negative sampling gradually gains popularity in temporal graph learning research~\citep{TGB} too, as single negative sampling may not be able to distinguish between various baselines.

\header{\bf Negative Sampling Strategies.} 	Previous studies~\citep{edgebank,TGB} leverage historical edges as negative samples to increase the difficulty for models in predicting potential links, based on the assumption that positive edges are likely repetitions of historical edges. However, this assumption does not hold in the domains covered by our TGB-Seq datasets, where historical edges are unlikely to reoccur in future time steps. Consequently, we randomly sample negative destination nodes from all possible nodes, specifically selecting from all nodes in non-bipartite datasets and all items in bipartite recommendation datasets.

\header{\bf Evaluation Metrics.} 
%In previous studies, Area Under the Receiver Operating Characteristic curve (AUROC) and Average Precision (AP) are commonly-adopted to evaluate link prediction performance. However, AUROC and AP are not proper metrics for link prediction with multiple negative samples as argued in~\citet{evaluating,TGB}.
Most existing studies leverage Area Under the Receiver Operating Characteristic curve (AUROC) and Average Precision (AP) for link prediction evaluation with a single negative sample, while ~\citet{evaluating}and~\citet{TGB} argue that they are not proper metrics for link prediction with multiple negative samples.
% AP is an appropriate metric for single negative samples, however, it cannot capture the ranking of a positive sample among multiple negative ones.  
Thus, we deploy the widely used ranking metric, Mean Reciprocal Rank (MRR) for evaluations, following~\citep{GraphMixer, TGB}. 
% The MRR score is defined as the average of the reciprocal ranks of the positive destination nodes among the negative destination nodes, and thus emphasizes the relative high likelihood of the positive edge among the candidate edges.
\begin{figure}[t]
  \begin{minipage}[t]{\textwidth}
  \centering
  % \vspace{1mm}
  \begin{tabular}{c}
  \hspace{-8mm} 
  \includegraphics[width=150mm]{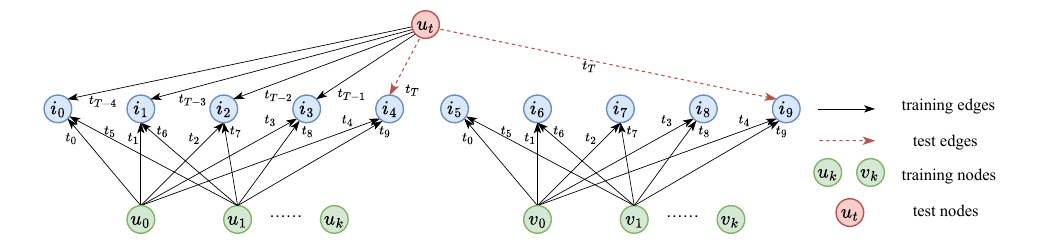}
  \end{tabular}
  \vspace{-2mm}
  \caption{Toy example of sequential dynamics in a temporal graph. The bipartite graph consists of users and items. The first user in group $u$, $u_0$, interacts sequentially with items $\{i_k\}_{k=0}^{k=4}$ at time $\{t_k\}_{k=0}^{k=4}$, respectively. Similarly, the first user in group $v$, $v_0$, interacts sequentially with items $\{i_k\}_{k=5}^{k=9}$ at the same timestamps $\{t_k\}_{k=0}^{k=4}$ as $u_0$. The second users, $u_1$ and $v_1$, follow a similar interaction pattern but interact with items at different times compared to the first users. All other users interact with items in a comparable sequential manner. A test sample queries whether the test node will interact with $i_4$ or $i_9$ at time $t_T$, based on its four previous interactions from $t_{T-4}$ to $t_{T-1}$.
   }
  \label{fig:example}
  \vspace{-4mm}
  \end{minipage}
\end{figure}

\subsection{Current Pitfalls in Temporal GNNs}
In this section, we aim to demonstrate that existing temporal GNNs are unable to capture even simple sequential dynamics. 
Figure~\ref{fig:example} illustrates a toy example of sequential dynamics in a temporal graph. To empirically evaluate whether existing temporal GNNs can learn the simple sequential dynamics, we construct a dataset that mirrors the dynamics depicted in Figure~\ref{fig:example}. Specifically, the dataset consists of items $\{i_k\}_{k=0}^{k=9}$ and multiple nodes in both group $u$ and group $v$, as in the toy example. To ensure that the sequential dynamics can be effectively modeled, the number of nodes in both group $u$ and group $v$ is set to 500. 
\begin{wraptable}{r}{0.3\textwidth}
  \centering
  \vspace{-5mm}
  \caption{The AP metric on the toy example dataset. $\ell$ indicates the length of the temporal random walk of CAWN. }
  \label{tbl:example}
  \vspace{2mm} 
  \resizebox{0.9\linewidth}{!}{
    \begin{tabular}{cc}
      \toprule
      Method & AP (\%)\\
       \midrule
      JODIE  & 51.19 ± 0.32  \\
      DyRep  & 51.30 ± 0.27  \\
      TGAT  & 51.06 ± 0.23  \\
      TGN  & 51.25 ± 0.48   \\
      CAWN ($\ell=1$) & 50.00 ± 0.00   \\
      CAWN ($\ell=2$) & 52.80 ± 0.05   \\
      EdgeBank & 50.00 ± 0.00 \\
      TCL & 50.00 ± 0.00\\
      GraphMixer & 50.00 ± 0.00  \\
      DyGFormer & 50.66 ± 0.50  \\  
      \midrule
      SGNN-HN & 100.00 ± 0.00 \\
      \bottomrule
      \end{tabular}
      \vspace{-35mm}
  }
\end{wraptable} 
Each $u_k$ interacts sequentially with items $\{i_k\}_{k=0}^{k=4}$, while each $v_k$ interacts sequentially with items $\{i_k\}_{k=5}^{k=9}$. Note that each $u_k$ and $v_k$ always interact at the same timestamps as stated in the caption of Figure~\ref{fig:example}. Both nodes and edges lack features.
%The node feature and edge feature are omitted.
The dataset is chronologically split into training set, validation set, and test set. The training set contains the complete interactions of 70\% of the users in both group $u$ and group $v$. 
% Both validation and test sets only contain the final interactions of 15\% nodes to ensure the sequential dynamics are exhibited multiple times in the training set. 
Given the four historical interactions, i.e, $\{i_k\}_{k=0}^{k=3}$ or $\{i_k\}_{k=5}^{k=8}$, a temporal GNN model is required to predict the interaction likelihood of the query user with $i_4$ and $i_9$.
%as the next item the query user will interact with.
Despite these straightforward sequential dynamics appearing commonly in the training set and thus considered as simple patterns, existing methods cannot correctly predict item $i_4$ instead of $i_9$ given that a test node has interacted with $\{i_k\}_{k=0}^{k=3}$ sequentially.
We use the AP metric to evaluate nine temporal GNNs and SGNN-HN. All temporal GNNs achieve an AP score of approximately 50\% as shown in Table~\ref{tbl:example}, indicating that they cannot distinguish between $i_4$ and $i_9$. 
% Detailed experimental settings are deferred to Appendix~\ref{app:toy}.

%In the following, we elaborate on the reasons that lead to the failure of existing temporal GNNs in capturing sequential dynamics. 
The shortcomings of current temporal GNNs in capturing sequential dynamics might relate to the functionality of their structures.
Generally, the temporal GNN models can be partitioned into two components: i) a memory module to represent the interaction history of the nodes, and ii) an aggregation module to aggregate neighborhood information when predicting future interactions. 
Among the existing studies, the designed temporal GNN models might contain both or either of these two components. The limitations of each component in capturing sequential dynamics to distinguish items $i_4$ and $i_9$ are discussed as follows.
% For those only with the memory module (Jodie and DyRep)~\citep{Jodie, DyRep}, they simply employ time projection and identity mapping when predicting future links. While for those only with the aggregation module, i.e., TCL, TGAT, CAWN, GraphMixer, and DyGFormer~\citep{TCL,TGAT,CAWN,GraphMixer,DyGFormer}, they \modif{XXX}. These methods might lack of the capacity for some aspects compared with containing both modules, e.g., TGN~\cite{}. 

%Some methods contains both modules, such as TGN, while others only contain one module, such as Jodie and DyRep only contain the memory module and simply employ time projection and identity mapping when predicting future links, respectively, and TCL, TGAT, CAWN, GraphMixer, and DyGFormer only contain the aggregation module. We discuss the limitations of these two modules in capturing sequential dynamics in the following.
\header{\bf Notations.} We denote the node feature of $u$ as $\nodef{u}$, the edge feature of $(u,v)$ as $\edgef{u}{v}$, and the time interval between the interaction $(u,v)$ and the query time as $\Delta t$. $\mem(u)$ and $\emb(u)$ denote the memory and embedding of node $u$, respectively. $\mathcal{N}^k_t(u)$ denotes the set of $k$-hop historical neighbors of node $u$ before time $t$. We use $\mathcal{N}_{\bf b}(u)$ to denote the set of $u$'s neighbors within a batch.

% Specifically, when a new edge arrives, the model updates the memories of the source and destination nodes based on their node feature, the edge feature, and the timestamp, to accurately represent the up-to-date interaction history of the nodes.

\header{\bf Memory module.} The memory module is designed to memorize the interaction history of nodes using a low-dimensional representation, called \textit{memory}. Formally, a node $s$'s memory is updated when processing a batch of incoming edges that involves $s$:
\begin{equation}
  \mem(s) = \memfunc\left(\mem(s), \nodef{s}, \left\{\left( \mem(d), \edgef{s}{d}, \Delta t \right) \mid d\in \mathcal{N}_{\bf b}(s) \right\}\right),
\end{equation}
where $\memfunc$ is typically an RNN model such as LSTM and GRU~\citep{GRU,LSTM}. The memory and feature of $s$, $\mem(s)$ and $\nodef{s}$, are treated as the hidden state of the RNN. The information of incoming edges in the batch, $\left\{\left( \mem(d), \edgef{s}{d}, \Delta t \right) \mid d\in \mathcal{N}_{\bf b}(s) \right\}$, serve as the input to RNN. Typically, only the last edge for each node in the batch is considered.
% The update strategies could be various, e.g., recurrent neural networks (RNNs)~\citep{RNN} and its variants~\citep{GRU,LSTM}.
%vary across methods, including recurrent neural networks (RNNs)~\citep{RNN} and its variants, GRUs~\citep{GRU} and LSTMs~\citep{LSTM}. 
%However, simply utilizing the information of historical edges makes the $k$-th users $u_k$ and $v_k$ share the same memory representation due to their similar interaction history, so does item $i_4$ and $i_9$. Therefore, the memory module cannot capture the difference between item $i_4$ and $i_9$. 
In the toy example, items $i_4$ and $i_9$ always interact at the same timestamps, resulting in undistinguishable memories.

%When predicting future edges $(s,d,t)$, the aggregation module aggregates the information from the historical neighbors of node $s$ and $d$ before time $t$ to generate their current embeddings. Common aggregation models includes mean-pooling, Transformer~\citep{attention} and its variants, and MLP-mixer~\citep{tolstikhin2021mlp}. 
\header{\bf Aggregation module with one-hop historical neighbors.} 
Given a prediction request for potential edges $(s,d,t)$, the aggregation module aggregates the information from the historical neighbors of node $s$ and $d$ before time $t$ to generate their current embeddings. The aggregation operation for node $s$ could be formulated as:
\begin{equation} \label{eq:agg}
  \emb(s) =\aggfunc\left( \nodef{s}, \left\{\left( \nodef{d}, \edgef{s}{d}, \Delta t \right) \mid d\in \mathcal{N}^k_t(s) \right\}\right),
\end{equation}
where $\aggfunc$ is commonly implemented as a Transformer~\citep{attention} (e.g., in DyGFormer) or its variants, an MLP-mixer~\citep{tolstikhin2021mlp} (e.g., in GraphMixer), or time projection function (e.g., in JODIE). Note that if the memory is available, $\nodef{s}$ is replaced by a combination of the memory and node feature.
In addition to interaction information, several studies compute the correlations between the neighborhoods of $s$ and $d$ to capture their structural and temporal dependencies:
\begin{equation}\label{eq:cor}
  \rel(s,d) = \CoNeifunc\left( \mathcal{N}^k_t(s), \mathcal{N}^k_t(d) \right),
\end{equation}
where $\CoNeifunc$ is the correlation function, such as calcu the number of common neighbors in DyGFormer and anonymous temporal random walk in CAWN. For computational efficiency, aggregation modules typically consider only one-hop neighbors. }

Such aggregation modules fail to capture the sequential dynamics in the toy example. The underlying issue is similar to that of memory modules: a node is represented solely by its null features and interaction timestamps. However, $i_4$ and $i_9$, their one-hop neighbors $\left\{ u_k \right\}$ and $\left\{ v_k \right\}$, interact in a similar manner at identical timestamps, respectively. As a result, the aggregation modules generate the same embeddings for $i_4$ and $i_9$, as well as for $\left\{ u_k \right\}$ and $\left\{ v_k \right\}$, respectively. While computing correlations between the source and destination nodes may seem helpful, both DyGFormer and CAWN fail in the toy example. DyGFormer's $\CoNeifunc$ is ineffective when the source and destination nodes have no common neighbors. Though CAWN's $\CoNeifunc$ employ a sophisticated anonymous random walk technique, it fails to distinguish between $i_4$ and $i_9$ because their one-hop neighborhoods mirror each other. Therefore, the aggregation modules of existing temporal GNNs cannot capture even the simple sequential dynamics in the toy example.
% As our toy example is a bipartite graph, where users and items do not share common neighbors, DyGFormer merely counts the number of historical interactions between the source and destination node, resulting in poor performance, as all users do not repeatedly interact with the same item. 

\header{\bf Aggregation module with high-order historical neighbors.} Leveraging high-order historical neighbor information can modestly enhance the capture of sequential dynamics. For example, extending the length of the temporal random walk from 1-hop to 2-hop in CAWN enables the incorporation of higher-order temporal and structural entangled information, resulting in a slight performance improvement from 50.00\% to 52.80\%.
The limited gain arises because an increased number of high-order neighbors introduces excessive noise. Consequently, CAWN is unable to effectively differentiate the subtle differences between the local structures of $(u_k, i_4)$ and $(u_k, i_9)$.
Furthermore, utilizing high-order information results in substantial computational resource consumption~\citep{HOT}. CAWN encounters memory issues on a GPU with 80GB of memory when the walk length is extended to three, even on this small graph. Therefore, effectively capturing intricate sequential dynamics through high-order neighbors remains an open problem.
%If we continue to increase the walk length, CAWN will run out of memory on 80G memory GPU for the toy example dataset.
% Thus, it is still an open problem to capture the correlations between source and destination nodes by high-order neighbors.
%The improvement is minor appears to be due to the high-order neighbors introduce excessive amounts of noise such that CAWN cannot effectively capture the subtle differences between the local structure of $(u_k,i_4)$ and $(u_k,i_9)$. 
%These results indicate that how to capture the correlations between source nodes and destination nodes by high-order neighbors information effectively is still an open problem.

In summary, neither the memory module nor the aggregation module can distinguish items $i_4$ and $i_9$ in the toy example. Consequently, temporal GNNs that incorporate either or both of these modules are unable to effectively capture the simple sequential dynamics, resulting in suboptimal performance on the toy dataset.
These findings suggest that current methods are insufficient for future link prediction that involve complex sequential dynamics and highlight the urgent need to establish new benchmark datasets to challenge temporal GNNs with sequential dynamics.
%These limitations emphasize the importance of new benchmark datasets that challenge existing methods with complex sequential dynamics and open up opportunities for future research on more effective and efficient temporal GNNs.

% \header{\bf Problem Definition}
% \header{\bf Memory-based methods}
% \paragraph{Stateless methods}

\section{Proposed Datasets}\label{sec:datasets}
Our proposed TGB-Seq aims to challenge temporal GNNs with intricate sequential dynamics that are inherently exhibited in various real-world dynamic systems. TGB-Seq comprises eight temporal graph datasets, including four bipartite datasets derived from recommender systems and four non-bipartite datasets curated from diverse application domains. All TGB-Seq datasets focus on interactions between entities and exclude node and edge features. Table~\ref{tbl:new-datasets} presents the statistics of TGB-Seq datasets.
% For bipartite graphs, we provide the number of users and items, while for non-bipartite graphs, we provide the total number of nodes. 
Besides, we also provide a selected list of datasets used for continuous-time temporal graph learning in Table~\ref{tbl:old-datasets} for comparison. %in Appendix. 

\begin{table}[t]
  \centering
  \caption{Statistics of TGB-Seq datasets.}
  \vspace{2mm}
  \label{tbl:new-datasets}
  \resizebox{\linewidth}{!}{
  % \begin{adjustbox}{max width=\textwidth}
  \begin{tabular}{lccccccc}
  \toprule
    \textbf{Dataset} & \textbf{Nodes (users/items)} & \textbf{Edges}  & \textbf{Timestamps} & \textbf{Repeat ratio(\%)} & \textbf{Density(\%)} &  \textbf{Bipartite} & \textbf{Domain}\\
    \midrule
   {ML-20M} & 100,785/9,646& 14,494,325& 9,993,250& 0& $1.49\times 10^{0}$& $\checkmark$ & Movie rating\\
    {Taobao}  &760,617/863,016& 18,853,792& 139,171& 16.58& $2.87\times 10^{-3}$& $\checkmark$ & E-commerce interaction\\
    {Yelp} & 1,338,688/405,081& 19,760,293& 14,646,734& 25.18& $3.64\times 10^{-3}$& $\checkmark$ & Business review\\
    {GoogleLocal} & 206,244/267,336& 1,913,967& 1,771,060& 0& $3.47\times 10^{-3}$&  $\checkmark$ & Business review\\
    {Flickr} & 233,836& 7,223,559& 134& 0& $1.32\times 10^{-2}$&   $\times$ & Who-To-Follow\\
    {YouTube} & 402,422& 3,288,028& 203& 0& $2.03\times 10^{-3}$&  $\times$ & Who-To-Follow\\
    {Patent} & 2,241,784& 12,749,824& 1,632& 0& $2.54\times 10^{-4}$&  $\times$ & Citation\\
    {WikiLink} & 1,361,972& 34,163,774& 2,198& 0& $1.84\times 10^{-3}$&  $\times$ & Web link\\
  \bottomrule
  \end{tabular}
  }
  \vspace{-5mm}
  % \end{adjustbox}
\end{table}

The most distinguishable feature of TGB-Seq datasets is the \textit{low repeat ratio}, where only the Yelp and Taobao datasets contain repeated edges due to the natural behavior of users who may click on items multiple times. The repeat ratio $r$ is defined as the portion of the number of repeated edges to the total number of edges in the dataset, i.e., 
$r= \frac{| \E_{\rm seen}|}{|\E|}$, where an edge $e_i=(s_i,d_i,t_i)\in\E_{\rm seen}$ if there exists an edge $e_j=(s_j,d_j,t_j)$ and satisfies that $s_i=s_j,d_i=d_j,t_j<t_i$. 

\header{\it Remark.} 
%Previous studies have observed the phenomenon that existing datasets exhibit excessive repeated edges and its impact on overly optimistic evaluations, and they focus on solving this issue with new evaluation protocols. 
The phenomenon of existing datasets that contain excessive repeated edges and its impact on overly optimistic evaluations has been highlighted in previous studies. To address the issues, these studies challenge the existing temporal GNNs with new evaluation protocols and new datasets from diverse domains. Specifically,
\citet{edgebank} proposes a historical negative sampling strategy to challenge existing methods with hard negative samples, and \citet{TGB} further employs multiple negative sampling strategies. Both of them propose new datasets from diverse domains and of diverse scales. However, most of the proposed datasets still contain numerous repeated edges as shown in Table~\ref{tbl:old-datasets}.
In contrast, we address this issue by proposing new challenging datasets curated to minimize repeated edges. Our TGB-Seq datasets emphasize intricate sequential dynamics, a key characteristic of many real-world applications. Consequently, TGB-Seq datasets provide a robust benchmark for evaluating the ability of temporal GNNs to capture sequential dynamics and generalize to unseen edges, a capability that is often lacking in existing benchmark datasets.
% ~\citet{edgebank} points out that randomly selected negative samples are \textit{easy} for temporal GNNs since edges tend to recur over time in existing datasets, such as Wikipedia and Reddit. Consequently, they propose a historical negative sampling strategy that selects historical edges as negative examples. Additionally, they introduce new datasets across diverse fields missing from current benchmarks. 
% Meanwhile, \citet{TGB} proposes large-scale benchmark datasets for future link prediction and employ a multiple negative sampling strategy that selects 20 negative destination nodes for each positive sample, prioritizing historical edges. 
% However, most of introduced datasets still exhibit a high ratio of repeated edges, as demonstrated in Table\ref{tbl:old-datasets}. 

% including two ``Who-To-Follow'' social networks, a citation network and a web link network. 

In addition to the low repeat ratio, another notable feature of TGB-Seq datasets is their origin in \textit{diverse domains that represent typical real-world applications of future link prediction}. 
% These include i) three recommendation scenarios: e-commerce interactions, movie ratings, and business reviews; ii) the “Who-To-Follow” task in social networks; iii) citation networks; and iv) web link networks. 
Besides classical applications like recommendations, the proposed non-bipartite datasets also represent fundamental applications in real-world contexts. As a crucial task for online social networking platforms, ``Who-To-Follow'' aims to recommend a list of users that a given user may be interested in following~\citep{whotofollow}. Effective prediction of relevant connections between users can significantly enhance user experience by fostering engagement and interaction within the platform. Moreover, future link prediction in citation networks and web link networks can be applied to knowledge graph completion, thereby enriching knowledge representations and enabling more comprehensive information retrieval~\citep{knowledgecompletion}.
Furthermore, TGB-Seq datasets exhibit several key attributes of real-world networks. Specifically, all TGB-Seq datasets adhere to a \textit{power-law degree distribution} (see Appendix~\ref{app:data}, Figure~\ref{fig:nodedegreedist}) and are notably sparse. Additionally, each dataset is \textit{medium to large scale}, containing millions to tens of millions of edges, which is consistent with the scale typically encountered in real-world networks.
% Moreover, the time granularity varies across TGB-Seq datasets, where all the recommendation datasets consist of unix-timestamps, while Flickr, YouTube, and WikiLink datasets are organized on a daily basis, and the patent datasets are weekly. 

% The low repeat ratio in TGB-Seq datasets compel existing methods to learn the intricate sequential dynamics and generalize to unseen edges, rather than simply relying on the historical edges. 
% We do not employ the \textit{surprise index} proposed in~\citet{edgebank} to capture the repeated level of datasets, i.e., $\frac{|\E_{\rm test}\backslash \E_{\rm train}|}{|\E_{\rm test}|}$, since the surprise index focuses on the repetitions of training edges in the test set. However, existing temporal GNNs leverage not only the historical edges in the training set but also in the test set for link prediction (the streaming setting as illustrated in~\citep{TGB}), the repeated pattern in the test set should be considered to measure the comprehensive repeated level in the whole dataset. Therefore, the repeat ratio is more suitable to meet our requirements than the surprise index. 

\header{\bf Dataset preprocessing.} We split the datasets chronologically into training, validation, and test sets with a ratio of 70\%/15\%/15\%. In the training sets, we retain only nodes with a degree of at least three. Besides, only nodes appearing in the training set are included in the validation and test sets. These settings are designed to mitigate the effects of cold-start nodes and the high sparsity of the datasets on the evaluation.
Descriptions of TGB-Seq datasets are shown below.

\underline{ML-20M }\footnote{https://grouplens.org/datasets/movielens/20m/} is a widely used benchmark dataset in recommendation research, derived from the MovieLens website. It contains movie rating data, where each record includes the rating score of a user, ranging from 1 to 5, for a specific movie along with the timestamp of the rating. While the ratings represent explicit feedback, we transform this data into implicit feedback for our analysis, following~\citet{NCF}. Consequently, the ML-20M network is represented as a bipartite graph where users and movies serve as nodes, and an edge represents a user's rating of a movie at a given time. The task of ML-20M and the following recommendation datasets is to predict whether a given user will interact with a given item at a given time.

\underline{Taobao} \footnote{https://tianchi.aliyun.com/dataset/dataDetail?dataId=649}~\citep{Taobao1,Taobao2,Taobao3} is a user behavior dataset derived from the e-commerce platform Taobao. It contains user click data on products from November 25, 2017, to December 3, 2017. The dataset is a bipartite graph where users and products are nodes, and an edge represents a user's click on a product at a given time.

\underline{Yelp} \footnote{https://www.yelp.com/dataset} is a business review dataset sourced from Yelp, a prominent platform for business recommendations, including restaurants, bars, and beauty salons. It contains user reviews of businesses from 2018 to 2022. The dataset is a bipartite graph where users and businesses are nodes, and an edge represents a user's review of a business at a given time.

\underline{GoogleLocal}~\citep{googlelocal1,googlelocal2} is a business review dataset derived from Google Maps, with a smaller scale compared to Yelp. It contains user reviews and ratings of local businesses. Following the settings for the ML-20M dataset, we treat these ratings as implicit feedback. Similar to the Yelp dataset, the GoogleLocal dataset is a bipartite graph where users and businesses are nodes, and an edge indicates a user's review of a business at a given time.

\underline{Flickr}~\citep{flickr} is a ``Who-To-Follow'' social network dataset derived from Flickr, a photo-sharing platform with social networking features. 
% Users can establish friendships with one another through the \textit{contact} feature. 
The dataset was crawled daily from November 2 to December 3, 2006, and from February 3 to March 18, 2007 by ~\citet{flickr}. It is estimated to represent 25\% of the entire Flickr network. The Flickr dataset is a non-bipartite graph where users are nodes, and an edge represents the friendship established between users at a given time. The task for the “Who-To-Follow” datasets, including Flickr and YouTube, is to predict whether a given user will follow another specified user at a particular time.

\underline{YouTube}~\citep{youtube} is another ``Who-To-Follow'' social network dataset derived from YouTube, a video-sharing platform that includes a user subscription network. Similar to Flickr, the YouTube dataset is a non-bipartite graph where users are nodes, and an edge indicates the subscription of a user to another user at a given time.

\underline{Patent}~\citep{patent} is a citation network dataset of U.S. patents, capturing the citation relationships between patents from 1963 to 1999. The dataset is organized as a non-bipartite graph where patents are nodes, and an edge represents a citation made by one patent to another at the time of publication. The task for the Patent dataset is to predict whether a given patent will cite another given patent, given several of their established citations.

\underline{WikiLink}~\citep{wikilink1,wikilink2,wikilink3} is a web link network dataset derived from Wikipedia, containing the hyperlink relationships between Wikipedia pages. This dataset is a non-bipartite graph, where pages are nodes and edges indicate hyperlinks established from one page to another at a given time. The task for WikiLink is to predict whether a given page will link to another given page at a given time.

\section{Experiments}\label{sec:exp}
In this section, we evaluate the performance of existing temporal GNNs on TGB-Seq datasets. The selected temporal GNN models includes JODIE~\citep{Jodie}, DyRep~\citep{DyRep}, TGAT~\citep{TGAT}, TGN~\citep{TGN}, CAWN~\citep{CAWN}, TCL~\citep{TCL}, GraphMixer~\citep{GraphMixer}, and DyGFormer~\citep{DyGFormer}. The descriptions of these methods are provided in Appendix~\ref{sec:baselines}. We employ the DyGLib~\cite{DyGFormer} framework to conduct the experiments. We limit the running time of each method to 48 hours and omit the methods that require more than 24 hours to finish one training epoch, which are denoted as OOT (out of time). Each result is the average of three runs with different random seeds with reported standard deviation. % is also reported. 

\header{\bf Implementation details.} We follow ~\cite{TGN} to set a relatively small batch size to ensure timely updates for the memory module. Specifically, we set the batch size to 200 for the GoogleLocal dataset across all methods. For larger datasets, however, a batch size of 200 is too small and would incur unacceptable training costs for most methods. %for most methods, 
Thus, we increase the batch size to 400 for all other datasets to accelerate the training process. Following DyGFormer, we use a learning rate of 0.0001 across all methods and datasets. 
A grid search is performed to tune the hyper-parameters of each method on the validation set. Detailed configurations are provided in Appendix~\ref{app:para_config}.
\begin{table}[t]
  \centering
  \vspace{-3mm}
  \caption{MRR of eight popular temporal GNN methods and SGNN-HN on four recommendation datasets and two previously established datasets (e.g., Wikipedia and Reddit). ``OOT'' denotes that the method failed to complete one epoch of training within 24 hours. The \textcolor{red}{first}, \textcolor{blue}{second}, and \textcolor{teal}{third} place rankings are highlighted accordingly.}
  \label{tbl:rec}
  \vspace{2mm}
  \resizebox{0.9\linewidth}{!}{
  \begin{tabular}{l|cccc|cc}
  \toprule
  Datasets & ML-20M & Taobao & Yelp & GoogleLocal & Wikipedia & Reddit\\
  \midrule
  JODIE & 21.16 ± 0.73 & \textcolor{teal}{48.36 ± 2.18} & \textcolor{red}{69.88 ± 0.31} & \textcolor{teal}{41.86 ± 1.49} &  76.94 ± 0.28& 77.92 ± 0.10
\\ 
  DyRep & 19.00 ± 1.69 & 40.03 ± 2.40 & 57.69 ± 1.05 & 37.73 ± 1.34 &  68.09 ± 1.45& 75.30 ± 0.30
\\ 
  TGAT & 10.47 ± 0.20 & OOT & OOT & 19.78 ± 0.24 &  72.42 ± 0.38& 76.69 ± 0.52
\\ 
  TGN & \textcolor{blue}{23.99 ± 0.20} & \textcolor{blue}{60.28 ± 0.54} & \textcolor{blue}{69.79 ± 0.24} & \textcolor{blue}{54.13 ± 1.97} &  81.16 ± 0.19& 79.82 ± 0.26
\\ 
  CAWN & 12.31 ± 0.02 & OOT & 25.71 ± 0.09 & 18.26 ± 0.02 &  \textcolor{red}{88.23 ± 0.33}& \textcolor{blue}{87.31 ± 0.32}
\\ 
%   EdgeBank & 1.82 ± 0.00& OOT & 9.77 ± 0.00& 1.96 ± 0.00&  78.10 ± 0.00& 78.08 ± 0.00
% \\ 
  TCL & 12.04 ± 0.02 & 31.55 ± 0.03 & 24.39 ± 0.67 & 18.30 ± 0.02 &  45.47 ± 3.48& 36.09 ± 2.10
\\ 
  GraphMixer & \textcolor{teal}{21.97 ± 0.17} & 31.54 ± 0.02 & 33.96 ± 0.19 & 21.31 ± 0.14 &  72.14 ± 0.80& 71.73 ± 0.32
\\ 
  DyGFormer & OOT & OOT & 21.68 ± 0.20 & 18.39 ± 0.02 & \textcolor{blue}{ 84.64 ± 0.43}& \textcolor{teal}{83.57 ± 1.42}\\ 
  \midrule
  SGNN-HN & \textcolor{red}{33.12 ± 0.01} & \textcolor{red}{68.58 ± 0.21} & \textcolor{teal}{69.34 ± 0.44 }& \textcolor{red}{62.88 ± 0.51}  &  \textcolor{teal}{83.83 ± 0.55} &  \textcolor{red}{89.01 ± 0.17}\\
  \bottomrule
  \end{tabular}
  }
  \vspace{-3mm}
  \end{table}

\subsection{Future Link Prediction Performance} 
\header{\bf Performance on recommendation datasets.} Table~\ref{tbl:rec} presents the results on four recommendation datasets, ML-20M, Taobao, Yelp, and GoogleLocal. 
Additionally, the results on two existing datasets, Wikipedia and Reddit, are included for comparison.
We observe that all existing temporal GNNs underperform on ML-20M, GoogleLocal and Taobao, with a large margin compared to SGNN-HN, one of the state-of-the-art methods for sequential recommendation. Notably, on the Yelp dataset, JODIE and TGN perform similarly to SGNN-HN and significantly outperform other temporal GNNs. This contrasts with the results on Wikipedia and Reddit, where memory-based models generally perform worse than CAWN and DyGFormer. 
In contrast, CAWN and DyGFormer do not achieve satisfactory performance across all recommendation datasets. 

These observations suggest that different temporal GNNs exhibit varying abilities in capturing sequential dynamics across datasets. Popular state-of-the-art methods like CAWN and DyGFormer tend to struggle with generalizing to unseen edges in real-world applications, excelling instead in capturing repetitive patterns. Memory-based methods, particularly JODIE and TGN, on the other hand, tend to perform better on datasets with fewer repeated edges. These findings highlight the need for more versatile methods capable of addressing the diverse challenges posed by different datasets.

\header{\bf Performance on non-bipartite datasets.} Table~\ref{tbl:non-rec} presents the results on four non-bipartite datasets, Flickr, YouTube, Patent, and WikiLink. The rankings of existing temporal GNNs vary even more significantly across these datasets. DyGFormer and GraphMixer outperform other methods on Flickr and YouTube, respectively, while JODIE and TGN perform best on Patent and WikiLink. Almost all methods rank in the top three in at least one dataset, which contrasts with the performance on recommendation datasets, where only JODIE and TGN perform well.
These results highlight the diverse capabilities of existing temporal GNNs and emphasize the need for more adaptable methods that can generalize across a wide range of datasets. The varying results presented in Table~\ref{tbl:rec} and Table~\ref{tbl:non-rec} underscore that the TGB-Seq datasets offer a more comprehensive evaluation of temporal GNNs' capabilities and introduce new challenges for future link prediction tasks.

\begin{table}[t]
  \centering
  \vspace{-3mm}
  \caption{MRR score of eight popular temporal GNNs on four non-bipartite datasets. The \textcolor{red}{first}, \textcolor{blue}{second}, and \textcolor{teal}{third} place rankings are highlighted accordingly.}
  \label{tbl:non-rec}
  \vspace{2mm}
  \resizebox{0.78\linewidth}{!}{
  \begin{tabular}{l|cccc}
  \toprule
  Datasets & Flickr & YouTube& Patent& WikiLink  \\
  \midrule
  JODIE & \textcolor{teal}{46.21 ± 0.83} & 41.67 ± 2.86 & \textcolor{red}{24.60 ± 0.38} & \textcolor{blue}{57.94 ± 1.33} \\ 
  DyRep & 38.04 ± 4.19 & 35.12 ± 4.13 & \textcolor{teal}{21.01 ± 1.14} & 42.63 ± 1.33 \\ 
  TGAT & 23.53 ± 3.35 & 43.56 ± 2.53 & 8.49 ± 0.18 & OOT   \\ 
  TGN & 46.03 ± 6.78 & \textcolor{blue}{55.16 ± 5.89} & \textcolor{blue}{22.83 ± 2.25} & \textcolor{red}{62.94 ± 2.16} \\ 
  CAWN & \textcolor{blue}{48.69 ± 6.08} & 47.55 ± 1.08 & 12.34 ± 0.47 & OOT   \\ 
  TCL & 40.00 ± 1.76 & \textcolor{teal}{50.17 ± 1.98} & 10.60 ± 1.75 & 43.02 ± 2.16 \\ 
  GraphMixer & 45.01 ± 0.08 & \textcolor{red}{58.87 ± 0.12} & 18.97 ± 2.54 & \textcolor{teal}{48.57 ± 0.02}\\ 
  DyGFormer & \textcolor{red}{49.58 ± 2.87} & 46.08 ± 3.44 & 14.20 ± 2.93 & OOT  \\
  \bottomrule
  \end{tabular}
  }
  % \vspace{-2mm}
\end{table}

\begin{figure}[t]
  \begin{minipage}[t]{1\textwidth}
  \centering
  % \vspace{-1mm}
  \begin{tabular}{c}
  % \hspace{-22mm} 
  \includegraphics[width=130mm]{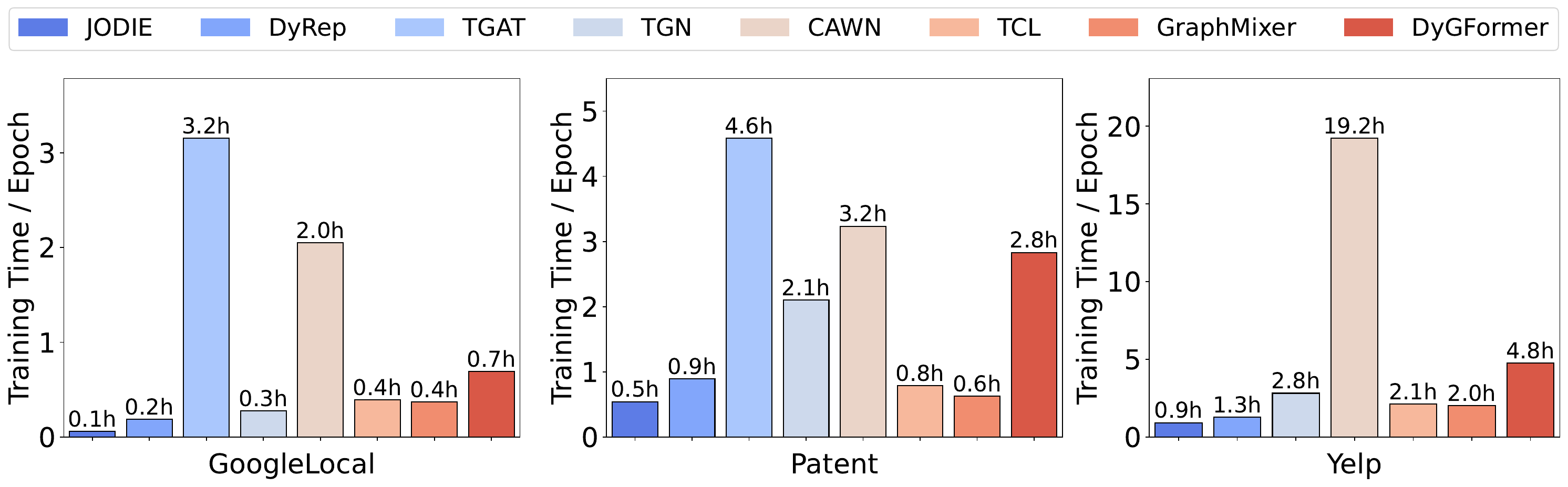}
  \end{tabular}
  \vspace{-4mm}
  \caption{The average training cost per epoch of eight popular temporal GNN methods on GoogleLocal, Patent, and Yelp datasets consists of 1.9M, 12.7M, and 19.7M edges, respectively.
  % The results of methods that fail to finish one training epoch in 24 hours are omitted.
   }
  \label{fig:training_cost}
  \vspace{-4mm}
  \end{minipage}
\end{figure}

\subsection{Training cost} 
To comprehensively study the efficiency of existing temporal GNNs, we select three datasets with various sizes of edge sets and report the average training cost per epoch of the corresponding approach.
Figure~\ref{fig:training_cost} illustrates the results on the GoogleLocal, Patent, and Yelp datasets,
where the methods that cannot finish one epoch in 24 hours are omitted. 
We observe that methods with simpler architectures, such as JODIE, DyRep, TCL and GraphMixer, exhibit significantly shorter training time compared to the others.
In contrast, TGAT and CAWN are the most inefficient methods, requiring considerable time to complete an epoch on the Patent and Yelp datasets. 
This inefficiency stems from the complex aggregation modules of these methods, which require multi-hop neighbor retrieval. While DyGFormer is more efficient than TGAT and CAWN on GoogleLocal, Patent, and Yelp, it still incurs higher costs. The calculation of co-occurrence frequencies for neighbors becomes particularly expensive when the temporal graph is dense. As demonstrated in Table~\ref{tbl:rec}, DyGFormer cannot complete an epoch within 24 hours for the ML-20M dataset, whereas other methods, including TGAT and CAWN, can.

These observations highlight that complex aggregation modules can significantly increase the training cost of existing temporal GNNs. 
As demonstrated in Table~\ref{tbl:rec} and Table~\ref{tbl:non-rec}, simpler methods like TCL and GraphMixer may be more efficient in terms of training, but they fail to achieve satisfactory performance. 
This investigation suggests that achieving both efficiency and effectiveness in temporal GNNs simultaneously remains an open problem, further underscoring the distinctive capability of TGB-Seq for comprehensive evaluations of these models.

\section{Conclusion}
In this paper, we demonstrate that current temporal GNNs struggle to capture intricate sequential dynamics inherent in real-world dynamic systems, thereby limiting their abilities to generalize across various future link prediction applications. However, existing datasets often contain excessively repeated edges and thus are inadequate for evaluating such abilities of temporal GNNs comprehensively. To address this gap, we propose TGB-Seq, a new challenging benchmark for temporal GNNs. 
TGB-Seq comprises eight datasets curated from diverse application domains characterized by complex sequential dynamics. Comprehensive evaluations on TGB-Seq reveal that existing temporal GNNs fail to achieve satisfactory performance across all datasets, underscoring the limitations of current methods and the necessity of TGB-Seq for robust temporal GNN evaluation.
% experience significant performance declines compared to their strong results on established benchmarks. This finding underscores the limitations of current methods' abilities in capturing intricate sequential dynamics and highlights the distinctive value of TGB-Seq in assessing these capabilities.

% \section*{Acknowledgements}
% The work was partially done at Gaoling School of Artificial Intelligence, Beijing Key Laboratory of Big Data Management and Analysis Methods, MOE Key Lab of Data Engineering and Knowledge Engineering, and Pazhou Laboratory (Huangpu), Guangzhou, Guangdong 510555, China. This research was supported in part by National Natural Science Foundation of China (No. 92470128, No. U2241212, No. 62276066), by National Science and Technology Major Project (2022ZD0114802), by Beijing Outstanding Young Scientist Program No.BJJWZYJH012019100020098, by Huawei-Renmin University joint program on Information Retrieval. We also wish to acknowledge the support provided by the fund for building world-class universities (disciplines) of Renmin University of China, by Engineering Research Center of Next-Generation Intelligent Search and Recommendation, Ministry of Education, by Intelligent Social Governance Interdisciplinary Platform, Major Innovation \& Planning Interdisciplinary Platform for the ``Double-First Class'' Initiative, Public Policy and Decision-making Research Lab, and Public Computing Cloud, Renmin University of China.

\bibliography{iclr2025_conference}
\bibliographystyle{iclr2025_conference}
\clearpage
\appendix
\begin{table}[t]
  \centering
  \caption{A selected list of datasets used for continuous-time temporal graph learning.}
  \label{tbl:old-datasets}
  \resizebox{\linewidth}{!}{
  % \begin{adjustbox}{max width=\textwidth}
  \begin{tabular}{lccccccc}
  \toprule
    \textbf{Dataset} & \textbf{Nodes (users/items)} & \textbf{Edges}  & \textbf{Timestamps} & \textbf{Repeat ratio(\%)} & \textbf{Density(\%)} &  \textbf{Bipartite} & \textbf{Domain}\\
    \midrule
   {ML-20M} & 100,785/9,646& 14,494,325& 9,993,250& 0& $1.49\times 10^{0}$& $\checkmark$ & Movie rating\\
    {Taobao}  &760,617/863,016& 18,853,792& 139,171& 16.58& $2.87\times 10^{-3}$& $\checkmark$ & E-commerce interaction\\
    {Yelp} & 1,338,688/405,081& 19,760,293& 14,646,734& 25.18& $3.64\times 10^{-3}$& $\checkmark$ & Business review\\
    {GoogleLocal} & 206,244/267,336& 1,913,967& 1,771,060& 0& $3.47\times 10^{-3}$&  $\checkmark$ & Business review\\
    {Flickr} & 233,836& 7,223,559& 134& 0& $1.32\times 10^{-2}$&   $\times$ & Who-To-Follow\\
    {YouTube} & 402,422& 3,288,028& 203& 0& $2.03\times 10^{-3}$&  $\times$ & Who-To-Follow\\
    {Patent} & 2,241,784& 12,749,824& 1,632& 0& $2.54\times 10^{-4}$&  $\times$ & Citation\\
    {WikiLink} & 1,361,972& 34,163,774& 2,198& 0& $1.84\times 10^{-3}$&  $\times$ & Web link\\
    \midrule
   {Wikipedia} & 8,227/1,000 &  157,474& 152,757& 88.41& $1.91\times 10^{0}$& $\checkmark$ & Interaction\\
    {Reddit}  & 10,000/984 & 672,447& 669,065& 88.32& $6.83\times 10^{0}$& $\checkmark$ & Social\\
    {MOOC} & 7,047/97 & 411,749& 345,600 & 56.66& $6.02\times 10^{1}$& $\checkmark$ & Interaction\\
    {LastFM} & 980/1,000 & 1,293,103 & 1,283,614& 88.01& $1.32\times 10^{2}$&  $\checkmark$ & Interaction\\
    {Enron} & 184& 125,235& 22,632& 90.79& $3.70\times 10^{2}$ & $\times$ & Social\\
    {Social Evo.} & 74& 2,099,519& 565,932& 99.77& $3.83\times 10^{4}$&  $\times$ & Proximity\\
    {UCI} & 1,899& 59,835& 58,911& 66.06& $1.66\times 10^{0}$&  $\times$ & Social\\
    {Flights} & 13,169& 1,927,145& 122& 79.50& $1.11\times 10^{0}$&  $\times$ & Transport\\
    {Contact} & 692& 2,426,279& 8,064& 96.72& $5.07\times 10^{2}$&  $\times$ & Proximity\\
    \midrule
    {tgbl-wiki} & 8,227/1,000 &  157,474& 152,757& 88.41& $1.91\times 10^{0}$& $\checkmark$ & Interaction\\
    {tgbl-review} & 352,636/298,590 & 4,873,540 & 6,865 & 0.19& $4.63\times 10^{-3}$&  $\checkmark$ & Rating\\
    {tgbl-coin} & 638,486 & 22,809,486 & 1,295,720 & 82.93& $5.60\times 10^{-3}$&  $\times$ & Transaction\\
    {tgbl-comment} & 994,790 & 44,314,507& 30,998,030 & 19.81& $4.48\times 10^{-3}$&  $\times$ & Social\\
    {tgbl-flight} &18,143 & 67,169,570& 1,385& 96.48& $2.04\times 10^{1}$&  $\times$ & Transport\\
    \midrule
    {Bitcoin-Alpha} &3,783& 24,186& 24,186 & 0& $1.69\times 10^{-1}$&  $\times$ & Finance\\
    {Bitcoin-OTC} &5,881& 35,592& 35,592&  0& $1.03\times 10^{-1}$&  $\times$ & Finance\\
    % {GDELT} & 16,682& 191,290,882& 170,522& & &  $\times$ & Event\\
  \bottomrule
  \end{tabular}
  }
  % \end{adjustbox}
\end{table}
\section{Dataset Documentation}
The resource links for TGB-Seq benchmark suits and datasets are provided as follows.
\begin{itemize}[leftmargin = *]
  \item The TGB-Seq website: \url{https://tgb-seq.github.io/}.
  \item The TGB-Seq datasets are available at Hugging Face: \url{https://huggingface.co/TGB-Seq}.
  \item The \texttt{tgb-seq} pip package is at \url{https://pypi.org/project/tgb-seq/}. The package will download datasets and negative samples automatically from Hugging Face.
  \item The TGB-Seq datasets can also be downloaded at \url{https://drive.google.com/drive/folders/1qoGtASTbYCO-bSWAzSqbSY2YgHr9hUhK}.
  \item The original datasets of TGB-Seq are available at \url{https://drive.google.com/file/d/1-1Ndp3R2qk_Jfk2zctReZIPOOzUuAQQI}.
\end{itemize}

\section{Further Information on TGB-Seq Datasets}\label{app:data}
\begin{figure}[t]
  \begin{minipage}[t]{1\textwidth}
  \centering
  % \vspace{1mm}
  \begin{tabular}{c}
  % \hspace{-22mm} 
  \includegraphics[width=140mm]{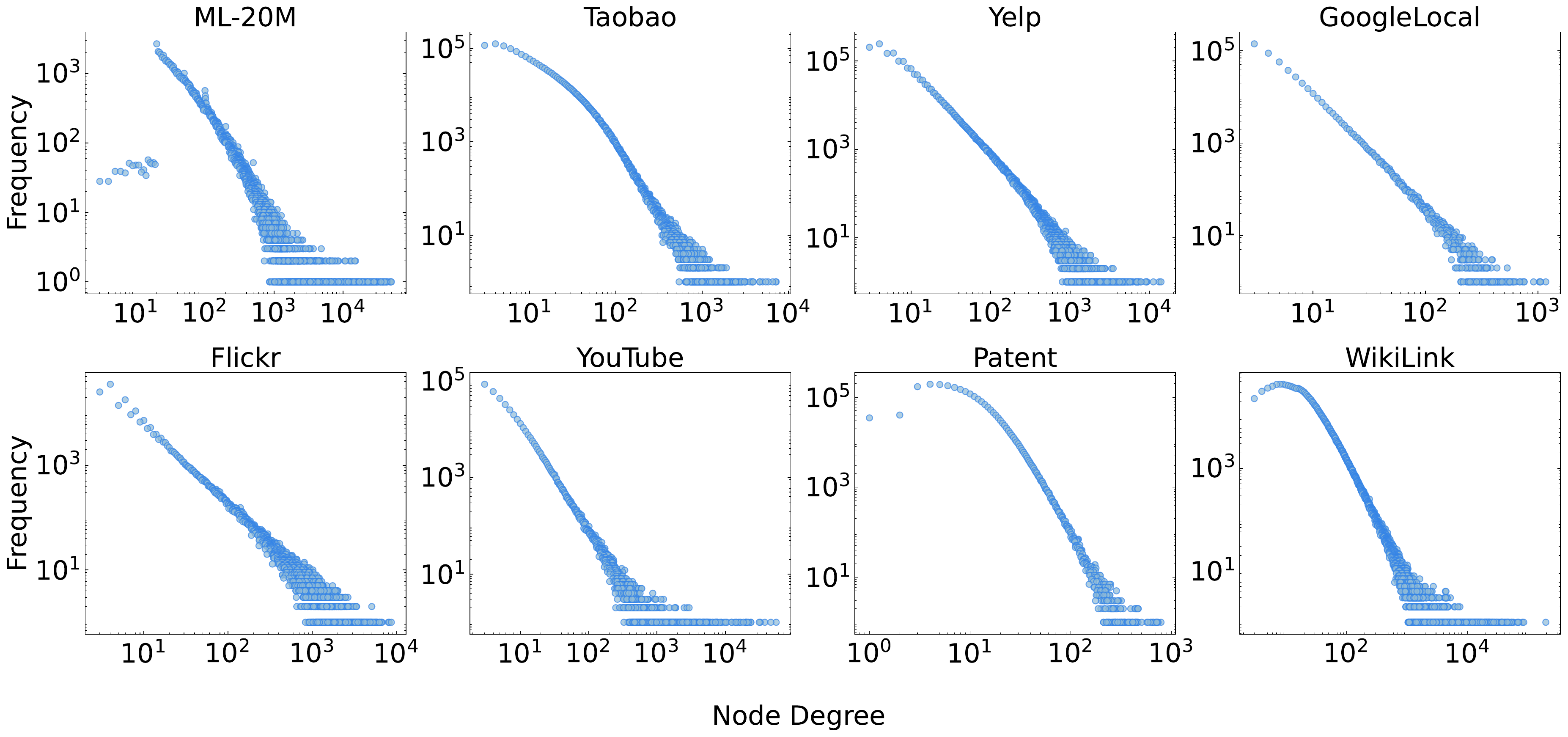}
  \end{tabular}
  \vspace{-4mm}
  \caption{Distribution of node degree on our TGB-Seq dataset.
   }
  \label{fig:nodedegreedist}
  \vspace{-4mm}
  \end{minipage}
\end{figure}

We provide a selected list of commonly used datasets for continuous-time temporal graph learning in Table~\ref{tbl:old-datasets} for reference. 
We also plot the node degree distribution of TGB-Seq datasets in Figure~\ref{fig:nodedegreedist}.
% We also present the variation in the number of incoming edges over discretized timestamps in TGB-Seq datasets, as shown in Figure~\ref{fig:edgedist}. Most datasets exhibit an increasing trend in the number of incoming edges as time progresses, while the Taobao dataset demonstrates periodic fluctuations, with phases of increase and decrease. These fluctuations are likely attributed to shopping festivals and other popularity-driven factors. The Flickr and YouTube datasets contain periods without any edges appearing due to the crawling process of the original datasets.
% The ML-20M, Yelp, Patent, and WikiLink datasets exhibit a continuous data distribution across their entire time span, indicating a stable trend. In contrast, the GoogleLocal, Flickr, and YouTube datasets show varying degrees of gaps in their data distribution, primarily due to the sampling process of the original datasets.
% Overall, the edge distributions of the eight datasets provide ample input data for various dynamic graph models, allowing for a comprehensive evaluation of their performance and efficiency.
Figure~\ref{fig:nodedegreedist} demonstrates that all the TGB-Seq datasets \textit{follow a power-law distribution}, a common characteristic of real-world networks~\citep{barabasi2013network}. This power-law behavior indicates that while a few hub nodes have a high degree of connections, the majority of nodes possess significantly fewer connections. Consequently, TGB-Seq is \textit{highly sparse}, exhibiting low density, as shown in Table~\ref{tbl:old-datasets}.

\header{\bf Preprocessing of the Patent dataset. } The Patent dataset is quite special that all citations of one patent are labeled with the same timestamp, specifically the publication time of the patent. Therefore, we carefully select test samples to ensure that each patent has prior citations, so that temporal GNNs are able to leverage these historical edges for future link prediction. Specifically, we choose not to validate or test the first 50\% of citations for the patents included in the validation and test sets; these citations serve solely as historical edges and are not used for model training. The remaining 50\% of citations are then evenly divided into validation and test samples. Although the citations of a patent occur simultaneously at the publication time, temporal GNNs can utilize the relative publication times of these patents and their neighbors to capture inherent research trends, thereby enhancing future link prediction performance.

\subsection{Node Features and Edge Features in TGB-Seq datasets}\label{sec:feat}
Most of TGB-Seq datasets, including all non-bipartite datasets and Taobao, do not include additional node features or edge features. However, other datasets, such as ML-20M and Yelp, include extra text features such as movies tags and businesses information. However, due to the absence of user-specific features, we cannot align the node features across users and items (i.e., movies and businesses in ML-20M and Yelp, respectively). The only dataset that includes additional text features for both items and users is GoogleLocal.
\begin{table}[t]
  \centering
  \caption{Performance comparison across models and various feature configurations on GoogleLocal.}
  \label{tab:feat}
  \resizebox{\linewidth}{!}{\begin{tabular}{lcccccccc}
  \toprule
  \textbf{}  & \textbf{JODIE} & \textbf{DyRep} & \textbf{TGAT} & \textbf{TGN} & \textbf{CAWN} & \textbf{TCL} & \textbf{GraphMixer} & \textbf{DyGFormer} \\ \midrule
  \textbf{w/o features} & 41.86 & {37.73} & 19.78 & 54.13 & \textbf{18.26} & \textbf{18.30} & 21.31 & 18.39   \\
  \textbf{+ edge features} & \textbf{42.44} & {37.22} & 17.35& 53.48 & 15.76 & 8.58& 21.38     & \textbf{18.53}    \\
  \textbf{+ node features} & 42.28 & \textbf{47.41} & \textbf{30.50} & \textbf{60.34} & 15.66 & 14.46  & \textbf{21.51}     & 17.91    \\
  \bottomrule
  \end{tabular}}
  \end{table}

In the following, we extract the text features from GoogleLocal and perform an empirical analysis of the impact of node and edge features on the performance of temporal GNNs. Specifically, the original GoogleLocal dataset contains node features for users and places, including \textit{name}, \textit{jobs}, \textit{currentPlace}, \textit{previousPlace}, \textit{education} for users, and \textit{name}, \textit{price}, \textit{address}, \textit{hour}, \textit{closed} for places. We treat \textit{price} and \textit{closed} as one-dimensional features, while the combination of other features is processed using SBERT~\citep{sbert} to generate semantic embeddings. These semantic embeddings are then reduced in dimensionality via PCA, resulting in a final embedding size of 172. Similarly, the edge features are processed as follows: the dataset includes user reviews of places, which consist of \textit{rating}, \textit{review\_text}, and \textit{category}. The \textit{rating} is treated as a one-dimensional feature, while \textit{review\_text} and \textit{category} are combined and processed using SBERT. The final dimensionality of edge features is also 172. SBERT is chosen for its ability to capture the semantic information of multilingual text, which is essential as the text in GoogleLocal is multilingual.

Table~\ref{tab:feat} demonstrates the performance of existing temporal GNNs on the GoogleLocal dataset when incorporating node features or edge features. The results indicate that the inclusion of node features significantly improves the performance of DyRep, TGAT, and TGN, while other methods show minimal improvement or even performance degradation when node or edge features are included. These findings suggest that while node features can enhance the performance of temporal GNNs, not all methods are equally effective at leveraging them for future link prediction tasks. This highlights the challenges of incorporating features into temporal GNNs and underscores the need for more robust and effective feature integration strategies in future research.

Additionally, it is important to note that the absence of features in future link prediction is a common scenario in real-world applications. A robust future link predictor must be able to capture the underlying dynamics based solely on interaction data. This is because features in real-world dynamic graphs are often incomplete, noisy, or difficult to align across different node types, especially in bipartite and heterogeneous graphs. For example, in the GoogleLocal dataset, user features and place features cannot be aligned due to their entirely different semantic meanings. Moreover, 267,200 out of 267,336 places in GoogleLocal lack the price feature entirely, further complicating the use of node features. These practical challenges explain the absence of features or the reliance on low-dimensional features in many existing temporal graph benchmarks. On the other hand, interaction data often plays a more critical role than feature data in link prediction tasks. For instance, SGNN-HN, which does not utilize any features, achieves the best performance among all temporal GNN models on GoogleLocal, including those that incorporate features. This highlights the pivotal importance of temporal interaction data in link prediction and emphasizes the significant progress that is still required for temporal graph learning models to fully harness the potential of such data.

% \begin{figure}[t]
%   \begin{minipage}[t]{1\textwidth}
%   \centering
%   % \vspace{1mm}
%   \begin{tabular}{c}
%   % \hspace{-22mm} 
%   \includegraphics[width=140mm]{./figs/edgestime_data.pdf}
%   \end{tabular}
%   \vspace{-4mm}
%   \caption{The variation of the number of edges in each discretized timestamp in the proposed TGB-Seq datasets.
%    }
%   \label{fig:edgedist}
%   % \vspace{-4mm}
%   \end{minipage}
% \end{figure}

\subsection{Dataset Licenses and Download Links}\label{app:datalicense}
In this section, we provide dataset licenses and download links as follows.

\underline{ML-20M}: The data set may be used for any research purposes under the following conditions:
(a)~The user may not state or imply any endorsement from the University of Minnesota or the GroupLens Research Group.
(b)~The user must acknowledge the use of the data set in publications resulting from the use of the data set.
(c)~The user may not redistribute the data without separate permission.
(d)~The user may not use this information for any commercial or revenue-bearing purposes without first obtaining permission from a faculty member of the GroupLens Research Project at the University of Minnesota.
(e)~The executable software scripts are provided "as is" without warranty of any kind, either expressed or implied, including, but not limited to, the implied warranties of merchantability and fitness for a particular purpose. 
The entire risk as to the quality and performance of them is with you.   Should the program prove defective, you assume the cost of all necessary servicing, repair or correction.
(f)~In no event shall the University of Minnesota, its affiliates or employees be liable to you for any damages arising out of the use or inability to use these programs (including but not limited to loss of data or data being rendered inaccurate). 
The original dataset can be found \href{https://grouplens.org/datasets/movielens/20m/}{here}.

\underline{Taobao}: CC BY-NC-SA 4.0 license (Creative Commons Attribution-NonCommercial-ShareAlike 4.0 International). 
The original dataset can be found \href{https://tianchi.aliyun.com/dataset/dataDetail?dataId=649}{here}.

\underline{Yelp}: MIT license. 
The original dataset can be found \href{https://www.yelp.com/dataset}{here}.

\underline{GoogleLocal}:
The original dataset can be found \href{http://jmcauley.ucsd.edu/data/googlelocal}{here}.

\underline{Flickr}: CC BY-SA license (Creative Commons Attribution-ShareAlike).
The original dataset can be found \href{https://networkrepository.com/soc-flickr-growth.php}{here}.

\underline{YouTube}: CC BY-SA license (Creative Commons Attribution-ShareAlike).
The original dataset can be found \href{https://networkrepository.com/soc-youtube-growth.php}{here}.

\underline{Patent}: MIT license.
The original dataset can be found \href{https://github.com/EdisonLeeeee/SpikeNet/tree/master}{here}.

\underline{WikiLink}: CC BY-SA license (Creative Commons Attribution-ShareAlike).
The original dataset can be found \href{https://networkrepository.com/web-wikipedia-growth.php}{here}.

\section{Further Discussion on Repeat and Exploration Behaviors in Temporal Graphs}\label{app:repeat}
We explore the future link prediction task in temporal graphs from the perspective of seen and unseen edges in Section~\ref{sec:intro}, which are analogous to repeat and exploration behaviors in recommendation systems.
Actually, repeat and exploration behaviors of users have been extensively studied in the context of recommender systems. Repeat behavior refers to users consistently engaging with items they have previously interacted with (i.e., the reoccurrence of seen edges in temporal graphs), while exploration behavior involves users discovering new items they have not interacted with before (i.e., the appearance of unseen edges for the first time in temporal graphs).
Existing studies reveal an imbalance in accuracy and difficulty between repetition and exploration in sequential recommendation tasks~\citep{RepetitionSR}. Several methods have been proposed to better address repeat and exploration behaviors, particularly in session-based or sequential recommendation~\citep{RepeatNet,Copy}, as well as next-basket recommendation~\citep{RepeatFood,masked}.
However, while repeat and exploration behaviors have been extensively studied in recommendation scenarios, their conclusions may not directly apply to the future link prediction task in temporal graphs due to differences in model design and task settings. For example, many recommendation methods are tailored for bipartite graphs without features or interaction timestamps, whereas temporal GNNs often focus on general graphs that may include single or multiple node types and fully leverage temporal graph information such as features and interaction timestamps.
Therefore, it is essential to investigate repeat and exploration behaviors in the context of future link prediction tasks on temporal graphs and to comprehensively evaluate the performance of existing temporal GNNs in handling these challenges. This is a key motivation behind the design of the TGB-Seq benchmark, which aims to provide a comprehensive evaluation of temporal GNNs across various datasets with diverse repeat and exploration behaviors.

\section{Experiments Details}
\subsection{Experimental Configurations}\label{app:para_config}
We perform a grid search to determine the optimal settings for key hyperparameters, with the search ranges and corresponding methods detailed in Table~\ref{tbl:grid search}. The final hyperparameter configurations identified through the grid search for various methods are summarized in Table~\ref{tbl:sampled neighbor} and Table~\ref{tbl:dropout}.
Regarding the configurations of neighbor sampling strategies, most methods achieve their best performance using the recent neighbor sampling strategy. However, for the ML-20M dataset, both CAWN and TCL achieve their best performance with the uniform neighbor sampling strategy.

For the ML-20M and the Flickr datasets, experiments are conducted on an Ubuntu machine equipped with Intel(R) Xeon(R) Gold 6240R CPU @ 2.40GHz. 
The GPU device is NVIDIA A100 with 80 GB memory. 
For the Taobao dataset, experiments are conducted on an Ubuntu machine equipped with Intel(R) Xeon(R) Gold 5218 CPU @ 2.30GHz. 
The GPU device is NVIDIA A100-SXM4 with 80 GB memory. 
For the Yelp dataset, experiments are conducted on an Ubuntu machine equipped with Intel(R) Xeon(R) Gold 6226R CPU @ 2.90GHz. 
The GPU device is NVIDIA RTX A6000 with 40 GB memory. 
For the GoogleLocal, the Patent, and the WikiLink datasets, experiments are conducted on an Ubuntu machine equipped with Hygon C86 7390 32-core Processor. 
The GPU device is NVIDIA A800 with 80 GB memory. 
For the YouTube dataset, experiments are conducted on an Ubuntu machine equipped with Intel(R) Xeon(R) Platinum 8369B CPU @ 2.90GHz. 
The GPU device is A100-SXM4 with 80 GB memory. 

\begin{table}[h]
\centering
\caption{Searched ranges of hyperparameters and the related methods.}
\label{tbl:grid search}
{
\begin{tabular}{c|cc}
\toprule
\textbf{Hyperparameters}  & \textbf{Searched Ranges}  & \textbf{Related Methods}   \\ 
\midrule
\begin{tabular}[c]{@{}c@{}}Number of \\ Sampled Neighbors\end{tabular}      & [20, 30, 40, 50, 60]     & \begin{tabular}[c]{@{}c@{}}DyRep, TGAT, TGN, CAWN, \\ TCL, GraphMixer\end{tabular}  \\

Dropout Rate   & \begin{tabular}[c]{@{}c@{}}[0.1, 0.3, 0.5]\end{tabular}  & \begin{tabular}[c]{@{}c@{}}JODIE, DyRep, TGAT, TGN, CAWN, \\ TCL, GraphMixer, DyGFormer\end{tabular} \\

\begin{tabular}[c]{@{}c@{}}Neighbor Sampling \\ Strategies\end{tabular}     & [recent, uniform] & \begin{tabular}[c]{@{}c@{}}DyRep, TGAT, TGN, CAWN,\\ TCL, GraphMixer\end{tabular} \\
\begin{tabular}[c]{@{}c@{}}Length of Input \\ Sequences \& \\ Patch Size\end{tabular} & \begin{tabular}[c]{@{}c@{}}[32 \& 1, 64 \& 2]\end{tabular}   & DyGFormer \\ 
\bottomrule
\end{tabular}
}
\end{table}

\begin{table}[h]
\centering
\caption{Configurations of the number of sampled neighbors and the length of input sequences $\&$ the patch size of different methods.}
\label{tbl:sampled neighbor}
{
\begin{tabular}{c|ccccccc}
\toprule
Datasets    & DyRep & TGAT & TGN & CAWN & TCL & GraphMixer & DyGFormer \\ 
\midrule
{ML-20M} & 20& 50& 40& 60& 60& 60&-\\
{Taobao}  & 20& -& 40& 60& 60& 60& -\\
{Yelp} & 20& -& 40& -& 60& 60& 32 \& 1\\
{GoogleLocal} & 20& 60& 40& 60& 60& 20& 64 \& 2\\
{Flickr} & 60& 40& 20& 40& 50& 40& 32 \& 1\\
{YouTube} & 20& 40& 60& 50& 40& 50& 32 \& 1\\
{Patent} & 60& 40& 60& 40& 40& 40& 64 \& 2\\
{WikiLink} & 40& -& 20& -& 60& 50& -\\ 
\bottomrule
\end{tabular}
}
\end{table}

\begin{table}[h]
\centering
\caption{Configurations of the dropout rate of different methods.}
\label{tbl:dropout}
{
\begin{tabular}{c|cccccccc}
\toprule
Datasets    & JODIE & DyRep & TGAT & TGN & CAWN & TCL & GraphMixer & DyGFormer \\ 
\midrule
{ML-20M} & 0.1& 0.1& 0.3& 0.1& 0.1& 0.1& 0.3& -\\
{Taobao}  & 0.1& 0.1& -& 0.1& 0.1& 0.1& 0.1& -\\
{Yelp} & 0.1& 0.1& -& 0.1& -& 0.1& 0.3& 0.1\\
{GoogleLocal} & 0.1& 0.3& 0.1& 0.1& 0.1& 0.1& 0.1& 0.1\\
{Flickr} & 0.1& 0.1& 0.1& 0.3& 0.1& 0.1& 0.1& 0.1\\
{YouTube} & 0.3& 0.1& 0.1& 0.1& 0.1& 0.1& 0.1& 0.1\\
{Patent} & 0.3& 0.3& 0.1& 0.1& 0.1& 0.1& 0.1& 0.1\\
{WikiLink} & 0.1& 0.1& -& 0.3& -& 0.1& 0.1& -\\ 
\bottomrule
\end{tabular}
}
\end{table}

% \begin{table}[h]
% \centering
% \caption{Configurations of neighbor sampling strategies of different methods.}
% \label{tbl:sample strategy}
% {
% \begin{tabular}{c|cccccc}
% \toprule
% Datasets    & DyRep & TGAT & TGN & CAWN & TCL & GraphMixer \\ 
% \midrule
% {ML-20M} & recent& recent& recent& uniform& uniform& recent\\
% {Taobao}  & recent& recent& recent& recent& recent& recent\\
% {Yelp} & recent& recent& recent& recent& recent& recent\\
% {GoogleLocal} & recent& recent& recent& recent& recent& recent\\
% {Flickr} & recent& recent& recent& recent& recent& recent\\
% {YouTube} & recent& recent& recent& recent& recent& recent\\
% {Patent} & recent& recent& recent& recent& recent& recent\\
% {WikiLink} & recent& recent& recent& recent& recent& recent\\ 
% \bottomrule
% \end{tabular}
% }
% \end{table}

\section{Temporal Graph Learning Methods}\label{sec:baselines}
We provide a brief overview of the temporal GNNs used in our experiments as follows.

\underline{JODIE}~\citep{Jodie} uses two coupled recurrent neural networks to dynamically update the states of users and items during interactions. 
It includes a novel projection operation that predicts future representation trajectories of both users and items, allowing the model to anticipate future behaviors. 
This architecture not only captures the evolution of user-item interactions but also facilitates the learning of representations that can be used for downstream tasks like recommendation and link prediction.

\underline{DyRep}~\citep{DyRep} introduces a dynamic representation learning framework that updates node states in real-time with each interaction. 
It leverages a recurrent neural network to capture node interactions and utilizes a temporal-attentive aggregation module to focus on evolving graph structures over time. 
DyRep is particularly effective in modeling dynamic relationships by considering both node communication and structural events, thus providing a comprehensive understanding of temporal graph changes.

\underline{TGAT}~\citep{TGAT} incorporates self-attention mechanisms to simultaneously model both the structural and temporal properties of dynamic graphs. 
Its design includes a time encoding function that uniquely represents temporal information, enabling the model to handle complex, evolving interactions among nodes. 
This combination allows TGAT to capture intricate temporal patterns and efficiently aggregate information from temporal-topological neighbors.

\underline{TGN}~\citep{TGN} introduces a memory-based approach for dynamic graph learning, where each node maintains an evolving memory that is updated through various interactions. 
Using a combination of message functions, aggregators, and memory updaters, TGN generates temporal node representations. 
The embedding module is crucial in capturing the temporal dynamics of nodes, which makes TGN adaptable for various dynamic graph tasks like link prediction and node classification.

\underline{CAWN}~\citep{CAWN} performs random walks on continuous-time dynamic graphs and employs an attention mechanism to selectively focus on crucial segments of these walks. 
This allows it to capture both temporal relationships and causal dependencies in the network. 
By learning these patterns, CAWN is capable of generating relative node identities, making it effective for temporal graph tasks such as anomaly detection and node classification.

\underline{EdgeBank}~\citep{edgebank} is a memory-centric approach tailored for transductive dynamic link prediction without relying on trainable parameters. 
It memorizes observed interactions and uses various strategies to update its memory. EdgeBank predicts future interactions based on whether the interaction is stored in its memory. 
Its simplicity lies in its rule-based decision-making, making it a lightweight yet competitive approach for link prediction in dynamic networks.

\underline{TCL}~\citep{TCL} employs contrastive learning on temporal graphs to learn robust node embeddings. 
Maximizing the agreement between node pairs that are temporally similar captures both temporal dependencies and topological structures. 
TCL uses a graph transformer to incorporate both graph topology and temporal information, along with cross-attention mechanisms to model interactions between nodes over time. 

\underline{GraphMixer}~\citep{GraphMixer} focuses on enhancing node embeddings in dynamic graphs by mixing both temporal and structural features. 
It uses a fixed time encoding function rather than a trainable one, incorporating it into a link encoder based on MLP-Mixer to learn temporal links effectively. 
GraphMixer also includes a node encoder with neighbor mean-pooling to aggregate node features, offering a comprehensive method for dynamic graph analysis.

\underline{DyGFormer}~\citep{DyGFormer} adopts a Transformer-based approach to capture long-term temporal dependencies in dynamic graphs. 
It introduces neighbor co-occurrence encoding and patching techniques, which help in modeling both the local and global structure of evolving interactions. 
This allows DyGFormer to effectively capture complex patterns in dynamic environments, making it suitable for various temporal graph tasks.

\end{document}